\documentclass[journal]{IEEEtran}
\usepackage{times}
\usepackage{psfrag,amsmath,amsfonts,verbatim,float,subfig}
\usepackage[demo]{graphicx}

\usepackage[numbers]{natbib}
\usepackage{url}
\usepackage{multicol}
\usepackage{float}
\usepackage[bookmarks=true, hidelinks]{hyperref}

\usepackage[short]{optidef}
\usepackage{algorithm}
\usepackage[noend]{algpseudocode}

\usepackage{booktabs}
\usepackage{bbding}
\usepackage{pifont}

\usepackage{pgfplots}
\usepackage{tikz}
\usepackage{tikzscale}
\usepgfplotslibrary{groupplots}
\usetikzlibrary{backgrounds}
\pgfplotsset{compat=1.15}
\def\checkmark{\tikz\fill[scale=0.4](0,.35) -- (.25,0) -- (1,.7) -- (.25,.15) -- cycle;}

\usepackage{xcolor}

\newcommand{\diff}[1]{\textcolor{black}{#1}}

\definecolor{orange}{RGB}{255,153,51}
\definecolor{cyan}{RGB}{51,255,255}
\definecolor{magenta}{RGB}{255,0,255}

\usepackage{soul}
\usepackage{caption}
\usepackage{tablefootnote}

\begin{document}
\title{Fast Contact-Implicit Model Predictive Control}
\makeatletter
\g@addto@macro\@maketitle{
\setcounter{figure}{0}
  \begin{figure}[H]
  \setlength{\linewidth}{\textwidth}
  \setlength{\hsize}{\textwidth}
  \centering
  \includegraphics[height=3.85cm]{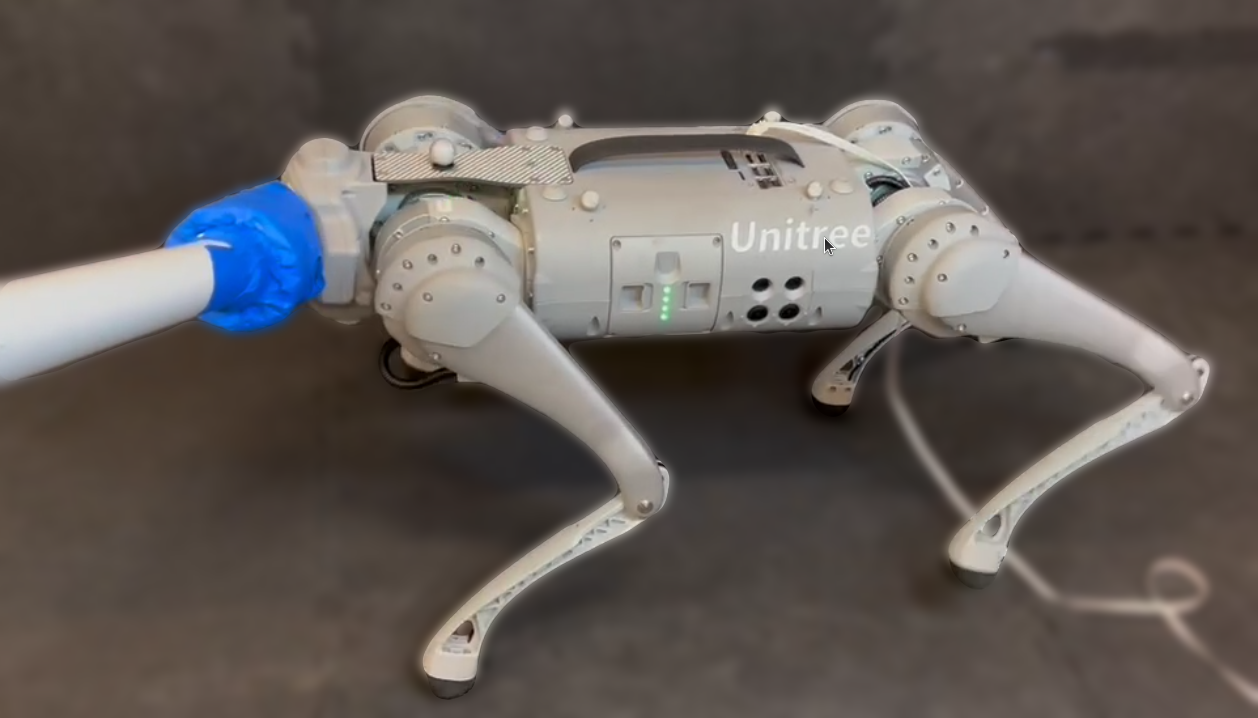}
  \hfill
  \includegraphics[height=3.85cm]{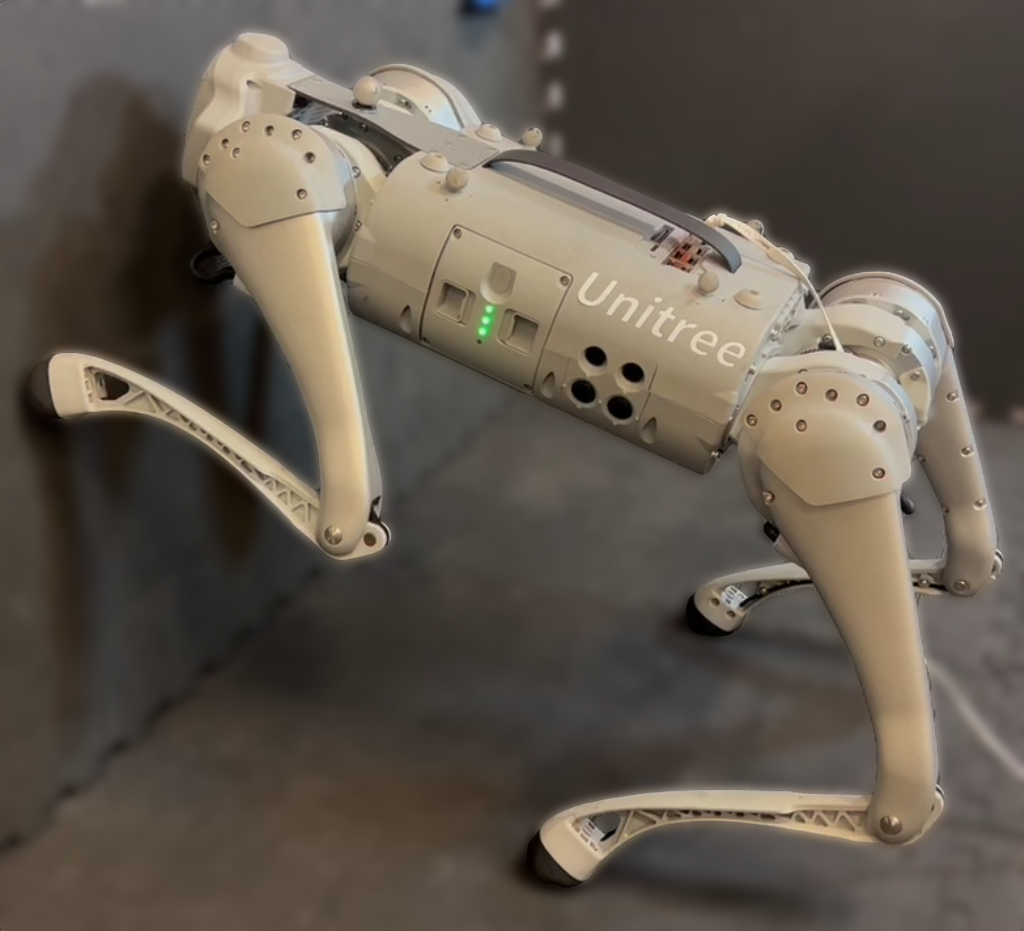}
  \hfill
  \includegraphics[height=3.85cm]{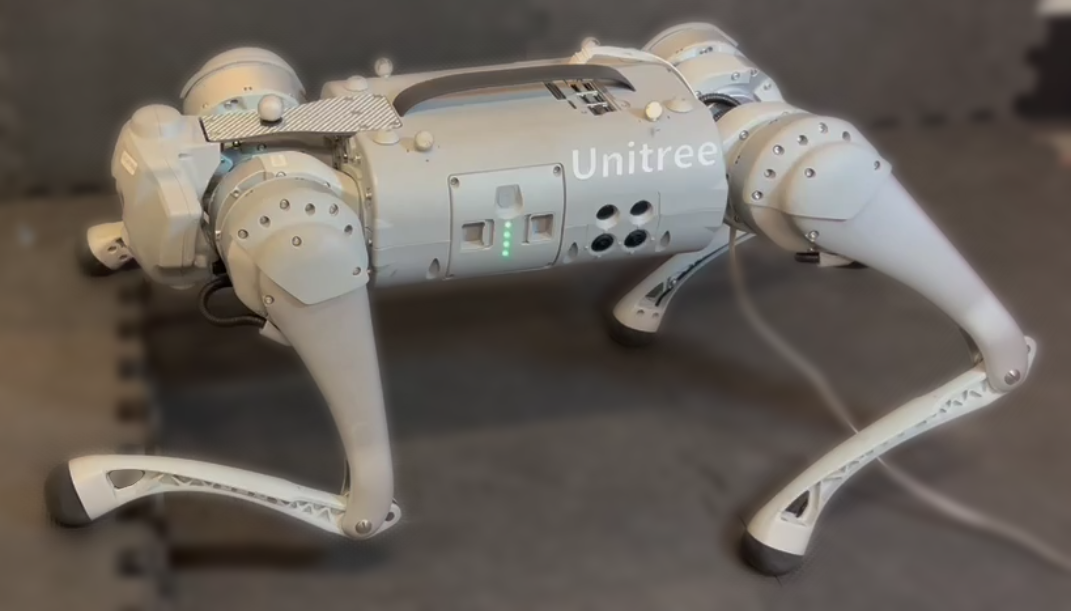}
  \caption{\diff{Hardware demonstrations on a Unitree Go1 quadruped: stable trotting while being pushed (left), transitioning from ground to standing against a wall (center), and placing two feet onto a step (right).}}
  \label{quadruped_hardware}
  \end{figure}
}
\makeatother

\author{Simon Le Cleac'h$^{1*}$, Taylor A. Howell$^{1*}$, Shuo Yang$^{2}$, Chi-Yen Lee$^{2}$, \\ John Zhang$^{2}$, Arun Bishop$^{2}$, Mac Schwager$^{3}$, and Zachary Manchester$^{2}$%
    \thanks{$^{1}$ Simon Le Cleac'h and Taylor A. Howell are with the Department of Mechanical Engineering, Stanford University,
		Stanford, CA 94305, USA.
		{\tt\footnotesize \{simonlc, thowell\}@stanford.edu}}%
    \thanks{$^{2}$ Shuo Yang, Chi-Yen Lee, John Zhang, Arun Bishop, and Zachary Manchester are with The Robotics Institute, Carnegie Mellon University,
		Pittsburgh, PA 15213, USA.
		{\tt\footnotesize \{shuoyang, chiyenl, johnzhang, arunleob, zacm\}@andrew.cmu.edu}}%
	\thanks{$^{3}$ Mac Schwager is with the Department of Aeronautics and Astronautics, Stanford University, 
		Stanford, CA 94305, USA.
        {\tt\footnotesize schwager@stanford.edu}}%
	\thanks{\textit{(Corresponding authors: Simon Le Cleac'h and Taylor Howell)}}
	\thanks{$^{*}$ These authors contributed equally to this work.}
}

\maketitle

\begin{abstract}
We present a general approach for controlling robotic systems that make and break contact with their environments. 
\diff{Contact-implicit model predictive control (CI-MPC) generalizes linear MPC to contact-rich settings by utilizing a bi-level planning formulation with lower-level contact dynamics formulated as time-varying linear complementarity problems (LCPs) computed using strategic Taylor approximations about a reference trajectory. 
These dynamics enable the upper-level planning problem to reason about contact timing and forces, and generate entirely new contact-mode sequences online.}
To achieve reliable and fast numerical convergence, we devise a structure-exploiting interior-point solver for these LCP contact dynamics and a custom trajectory optimizer for the tracking problem. 
\diff{We demonstrate real-time solution rates for CI-MPC and the ability to generate and track non-periodic behaviours in hardware experiments on a quadrupedal robot.} We also show that the controller is robust to model mismatch and can respond to disturbances by discovering and exploiting new contact modes across a variety of robotic systems in simulation, including a pushbot, planar hopper, planar quadruped, and planar biped.
\end{abstract}


\section{Introduction}
\IEEEPARstart{C}{ontrolling} systems that make and break contact with their environments is one of the grand challenges in robotics. Numerous approaches have been employed for controlling such systems, ranging from hybrid-zero dynamics \cite{westervelt2003hybrid, ames2014rapidly, li2021reinforcement}, to complementarity controllers \cite{aydinoglu2020contact}, to neural-network policies \cite{heess2017emergence,heiden2020neuralsim}, and model predictive control (MPC) \cite{winkler2018gait, sleiman2021unified}. There have also been numerous successes deploying such approaches on complex systems in recent years: direct trajectory optimization and LQR on Atlas \cite{kuindersma2016optimization}, smooth-contact models and differential dynamic programming on HRP-2  \cite{todorov2012mujoco, tassa2012synthesis, koenemann2015whole}, zero-moment point and feedback linearization on ASIMO \cite{hirai1998development}, and MPC with simplified dynamics models on Cheetah \cite{bledt2020regularized} and ANYmal \cite{lee2020learning}. However, reliable \emph{general-purpose} control techniques that can reason about contact events and can be applied across a wide range of robotic systems without requiring application-specific model simplifications, gait-generation heuristics, or extensive parameter tuning remain elusive.

\diff{Our approach combines fast, differentiable rigid-body dynamics with contact, strategic approximations about a reference trajectory, and specialized numerical optimization techniques for the application of local tracking control for systems that experience contact interactions with their environments. The result is a bi-level model predictive control algorithm that can effectively reason about contact changes in the presence of large disturbances while remaining fast enough for real-time execution.}

\diff{We formulate contact dynamics as a complementarity problem and devise a fast interior-point solver to reliably optimize this feasibility problem. Smooth gradients are efficiently computed through the non-smooth dynamics by exploiting intermediate solutions from within this solver using implicit differentiation.} To enable real-time performance for control, we pre-compute linearizations of the system's dynamics, signed-distance functions, and friction cones about a reference trajectory, while explicitly retaining complementarity constraints that encode contact switching behavior, resulting in a sequence of lower-level time-varying linear-complementarity problems (LCP) which represent the model's contact dynamics. An upper-level trajectory optimization problem is then optimized using fast linear algebra. We refer to this algorithm as \textit{Contact-Implicit Model Predictive Control} (CI-MPC).

\diff{Finally, we demonstrate that CI-MPC can generate new contact sequences online and reliably track reference trajectories despite significant model mismatch and while large external disturbances are applied for a number of qualitatively different robotic systems, including: a pushbot, and planar hopper, quadruped, and biped systems in simulation; and on Unitree Go1 quadruped hardware \cite{unitree_go1}.}

Our contributions are:
\begin{itemize}
	\item \diff{Fast approximate contact dynamics that can be reliably evaluated and efficiently differentiated with a custom interior-point solver}
	\item Structure-exploiting solvers for the contact-dynamics and trajectory optimization problems
	\item A model predictive control framework for robotic systems with contact dynamics
	\item \diff{A collection of simulation and hardware experiments demonstrating the performance of CI-MPC on a variety of robotic systems across a range of highly dynamic tasks}
	\end{itemize}

\diff{In the remainder of this paper, we first review related work on control through contact with MPC, as well as complementarity-based contact dynamics in Section \ref{related_work}. Next, we present a brief overview of MPC, outline the classic complementarity formulation for contact dynamics, and provide background on interior-point methods and implicit differentiation in Section \ref{background}. Then, we present CI-MPC in Section \ref{cimpc}. Results are presented in Section \ref{results} including both simulation and hardware experiments. Finally, we discuss our results, limitations of this approach, potential directions for future work in Section \ref{conclusion}.}

\section{Related Work} \label{related_work}
In this section, we review related work on MPC for the control of dynamical systems that make and break contact with their environments and provide an overview of complementarity-based contact dynamics.

\subsection{Model Predictive Control}

Today, most successful approaches for controlling legged robots utilize MPC in combination with simplified models and heuristics originally pioneered by Raibert for hopping robots \cite{raibert1989dynamically}. The key insight of this work is that the control problem can be decoupled into a high-level controller that plans body motions while ignoring the details of the leg dynamics, and a low-level controller that determines the necessary leg motions and joint torques to generate the forces and torques on the body determined by the high-level controller.

Arguably the most impressive control work on humanoids has utilized centroidal dynamics with full kinematics to enable Atlas to navigate various scenarios with obstacles \cite{dai2014whole} and perform parkour \cite{atlas2019parkour}. Integrating hardware design and controller synthesis has also recently enabled small humanoids to perform agile acrobatic maneuvers in simulation \cite{chignoli2021humanoid}.

There have also been impressive advances for quadrupeds, achieved by designing hardware that aims to closely match the modeling approximations made in the controller, e.g., building very light legs \cite{bledt2020regularized}. Whole-body control, which has the benefit of simpler overall control structures and the ability to leverage a system's dynamics, has been achieved at real-time rates on hardware \cite{neunert2018whole}. Approaches that utilize both force-based MPC and whole-body control have also demonstrated agile locomotion \cite{kim2019highly}.

A major limitation of these prior works is that the control policies are highly specialized to a specific robotic system. In this work, we compare CI-MPC to a number of system-specific control methods that perform quite well for their given system, but do not generalize to other systems, whereas our policy generalizes to many different systems that experience contact interactions while achieving comparable or better performance.

\subsection{Complementarity-Based Contact Dynamics}\label{lcp_dynamics}
\diff{The classic approach for simulating rigid-body dynamics with contact interactions is a velocity-based time-stepping scheme formulated as a linear complementarity problem (LCP). The LCP searches for the next state of the system while enforcing impact and friction constraints. Solvers for this class of problems utilize pivoting methods \cite{drumwright2015rapidly}, such as Lemke's algorithm \cite{cottle2009linear}, or interior-point methods \cite{kojima1991unified}. Implementations of pivoting methods can be found in general-purpose LCP solvers such as PATH \cite{dirkse1995path}, or physics engines including: Bullet \cite{coumans2019} and DART \cite{lee2018dart}.}

\diff{Derivatives of LCP-based contact dynamics can be efficiently computing using implicit differentiation \cite{dini1907lezioni}. However, the quality of these results is dependent on the method employed for optimization. Pivoting approaches enforce strict complementarity at each iteration, returning solutions at non-differentiable points. As a consequence, this differentiation will return subgradients that make, typically efficient, second-order optimization slower and less reliable. In contrast, interior-point methods relax the complementarity constraints at each iteration, only converging in the limit. These intermediate results can be implicitly differentiated to return smooth gradients \cite{howell2022dojo}. Alternative approaches for computing gradients for contact dynamics include utilizing auto-differentiation tools \cite{heiden2020neuralsim} and analyzing the LCP solution to select subgradients \cite{werling2021fast}.}

\diff{In addition to simulation, contact dynamics represented as LCPs have been utilized for planning. Collocation approaches \cite{stryk1993numerical} directly encode the LCP problem as  as constraints in order to enforce contact dynamics, along with an objective specifying desired behavior, in a large non-convex problem \cite{posa2014direct}.}
This approach enables the optimizer to plan without pre-specified mode sequences for locomotion and simple manipulation tasks. Subsequent work improved this approach by introducing higher-order integrators for the dynamics and a numerically robust, exact $\ell_1$-penalty for handling the complementarity constraints \cite{manchester2020variational}. \diff{Alternative rollout-based approaches utilize LCPs for forward simulation and subsequently differentiate through the solution of one-step dynamics in order to compute derivatives for gradient-based optimization \cite{de2018end,werling2021fast}.}

Another popular contact-dynamics formulation is MuJoCo's soft-contact model \cite{todorov2012mujoco}, which solves a convex optimization problem and trades physical realism for fast and reliable performance. An alternative model solved a strictly convex quadratic program \cite{anitescu2006optimization}. 
\diff{Gradients are computed use a finite-difference scheme. However, this approach is computationally less efficient.}
Additionally, the LCP complementarity constraints can be relaxed, resulting in a soft-contact model that exhibits improved numerical properties in some scenarios \cite{geilinger2020add}. 

\section{Background} \label{background}
\diff{In this section, we provide technical background on MPC, complementarity-based contact dynamics, interior-point methods, and implicit differentiation.}

\subsection{Model Predictive Control} \label{mpc_background}
\diff{Predictive control policies \cite{richalet1978model} optimize a planning problem:}
\begin{equation}
	\begin{array}{ll}
	\underset{x_{1:T}, u_{1:T-1}}{\mbox{minimize}} & g_T(x_T) + \sum \limits_{t = 1}^{T-1} g_t(x_t, u_t) \label{trajectory_optimization}\\
	\mbox{subject to} & x_{t+1} = f_t(x_t,u_t), \quad t = 1,\dots,T-1,\\
	& (x_1~\mbox{given}),
	\end{array}
\end{equation}
\diff{for a given initial state in order to compute controls for a dynamical system we aim to control. If planning is performed at a sufficiently high rate, the sequence of open-loop plans provide \textit{feedback}.} For a system with state $x \in \mathbf{R}^{n}$, control $u \in \mathbf{R}^{m}$, time index $t$, initial state $x_1$, and discrete-time dynamics $f : \mathbf{R}^{n} \times \mathbf{R}^{m} \rightarrow \mathbf{R}^{n}$, the optimizer aims to minimize an objective with costs, $g: \mathbf{R}^{n} \times \mathbf{R}^{m} \rightarrow \mathbf{R}$, over a planning horizon $T$.

\diff{Solving a (potentially) non-convex problem \eqref{trajectory_optimization} online can be unreliable or computation too expensive. Instead, a proxy problem is solved online that makes strategic approximations. A common simplification is to track a reference trajectory, $\bar{\tau} = (\bar{x}_1, \bar{u}_1, \dots, \bar{x}_T$), denoted with an overbar ($\bar{\phantom{a}}$), that is precomputed offline. In this setting, the computational complexity for planning is reduced by utilizing dynamics:
\begin{equation}
    \delta x_{t+1} = A_t \delta x_t + B_t \delta u_t, \label{linear_dynamics}
\end{equation}
linearized about the reference, where $A = \partial f(\bar{x}, \bar{u}) / \partial x$, $B = \partial f(\bar{x}, \bar{u}) / \partial x$, and the decision variables are relative to the reference trajectory, i.e., $\delta a = a - \bar{a}$; and a quadratic objective:
\begin{equation}
\frac{1}{2}\delta x_t^T Q_t \delta x_t + q_t^T \delta x_t + \frac{1}{2}\delta u_t^T R_t \delta u_t  + r_t^T \delta u_t, \label{quadratic_objective}
\end{equation}
where $Q = \partial^2 g(\bar{x}, \bar{u}) / \partial x^2, q = \partial g(\bar{x}, \bar{u}) / \partial x, R = \partial^2 g(\bar{x}, \bar{u}) / \partial u^2, r = \partial g(\bar{x}, \bar{u}) / \partial u$, similarly comprise a second-order expansion about the reference trajectory.}

\diff{This formulation, potentially with additional affine state and control constraints, is commonly referred to as \textit{linear} MPC. Without such constraints, the problem 
(\ref{trajectory_optimization}) with linear dynamics \eqref{linear_dynamics} and quadratic objective \eqref{quadratic_objective} is the LQR problem \cite{kalman1964lqr} and is efficiently solved with a backward Riccati recursion.}

\diff{Predictive control iteratively optimizes \eqref{trajectory_optimization}, or an approximate version of it (\ref{linear_dynamics} - \ref{quadratic_objective}), for a given state, and the optimized controls are utilized to compute an input for the system.} After the system evolves, the problem is re-optimized for a new state in order to compute a new control for the system. By repeating this procedure at a high rate, feedback control is achieved \cite{wang2009fast}. In practice, applying the controls optimized with time-varying linearized dynamics and quadratic costs, to the actual nonlinear system is extremely effective, especially for applications that track a reference trajectory.

\subsection{\diff{Complementarity-Based Contact Dynamics}}

\diff{Contact dynamics can be simulated with a velocity time-stepping scheme \cite{stewart1996implicit}, formulated as complementarity problem \cite{cottle2009linear}: 
\begin{align}
	\label{feas_prob}
	{\mbox{find}} \quad & q, \lambda,\eta, \beta, \psi \\
	\mbox{subject to} \quad & \left[M_{+}(q - q_{-}) - M_{-}(q_{-} - q_{--})\right] / h + h C = \notag\\ 
	& \quad J^T \lambda + B u, \label{smooth_dynamics}\\
	& \gamma \circ \phi = 0, \label{impact_complementarity}\\
	& \beta \circ \left[ P (q - q_{-})/h + \psi \textbf{1} \right] = 0, \label{friction_complementarity} \\
	& \psi \circ \left[\mu \gamma - \textbf{1}^T \beta \right] = 0, \label{friction_velocity_complementarity}\\
	& \phi, \gamma \geq 0, \label{impact_inequalities} \\
	& \beta, \psi, [P (q - q_{-})/h + \psi \textbf{1}], [\mu \gamma - \textbf{1}^T \beta] \geq 0, \label{friction_inequalities}
\end{align}
that finds the next configuration of the system $q \in \mathbf{R}^{n_q}$ using implicitly defined velocities, $v = (q - q_{-}) / h \in \mathbf{R}^{n_v}$. Subscripts indicate a previous time step. This formulation considers a single contact point, but generalizes to systems with multiple contacts.
The problem utilizes: the mass matrix $M: \mathbf{R}^{n_q} \rightarrow \mathbf{S}_{++}^{n_v}$; dynamics bias $C: \mathbf{R}^{n_q} \times \mathbf{R}^{n_v} \rightarrow \mathbf{R}^{n_v}$ that includes Coriolis and gravitational terms; contact Jacobian $J: \mathbf{R}^{n_q} \rightarrow \mathbf{R}^{d \times n_v}$ that maps contact forces in the contact frame into the generalized coordinates; input Jacobian $B: \mathbf{R}^{n_q} \rightarrow \mathbf{R}^{n_v \times m}$ that maps control inputs, typically joint torques, into the generalized coordinates; $P: \mathbf{R}^{n_q} \rightarrow \mathbf{R}^{p (d-1) \times n_v}$ is a mapping from the generalized velocity space to an overparameterized contact tangent space; time step $h \in \mathbf{R}_{+}$; contact forces $\lambda = (\gamma, \beta) \in \mathbf{R}^d$ defined in the contact frame, with normal force $\gamma \in \mathbf{R}$ and overparameterized friction forces $\beta \in \mathbf{R}^{p (d - 1)}$ that are constrained by a linearized friction cone; $\psi \in \mathbf{R}$ is a dual variable associated with friction, representing the magnitude of the contact point velocity;} signed-distance function, $\phi: \mathbf{R}^{n_q} \rightarrow \mathbf{R}$, that returns distance between a specified contact point on the robot (e.g., feet) and the closest surface in the environment (e.g., the floor); and where $\circ$ is an element-wise (Hadamard) vector product. \diff{We use $p$ to denote the overparameterization dimension (often $p$ = 2) and $d$ to denote the environment dimension $d = 2$ for planar systems and $d = 3$ otherwise.}

\diff{The smooth dynamics \eqref{smooth_dynamics} are discretized with a semi-implicit Euler scheme \cite{marsden2001discrete}; complementarity constraints (\ref{impact_complementarity}- \ref{friction_velocity_complementarity}) encode contact switching behavior; impact is encoded in \eqref{impact_complementarity} and \eqref{impact_inequalities}; and friction terms (\ref{friction_complementarity}-\ref{friction_velocity_complementarity}) and \eqref{friction_inequalities}, are derived from the maximum dissipation principle \cite{moreau2011unilateral}.}

\diff{Problem data include previous configurations $q_{-}, q_{--}$, time step $h$ and control inputs $u$. A nonlinear formulation uses the following mappings:
\begin{align}
    &M_{+} \leftarrow M(q), 
    &&M_{-} \leftarrow M(q_{-}), \notag \\ 
    &C \leftarrow C(q, (q - q_{-}) / h), 
    &&J \leftarrow J(q),  \\ 
    &B \leftarrow B(q), 
    && P \leftarrow P(q), \notag \\ 
    & \phi \leftarrow \phi(q), \notag
\end{align}
evaluating these terms at the next configuration. In practice, a partial linearization of the dynamics is performed to satisfy a linear complementarity problem (LCP) \cite{stewart1996implicit}, resulting in the following mappings:
\begin{align}
    &M_{+} \leftarrow M(q_{-}), 
    &&M_{-} \leftarrow M(q_{--}), \notag \\ 
    &C \leftarrow C(q_{-}, (q_{-} - q_{--}) / h), 
    &&J \leftarrow J(q_{-}),  \\ 
    &B \leftarrow B(q_{-}), 
    && P \leftarrow P(q_{-}), \notag \\
    & \phi \leftarrow \phi(q_{-}) + N(q{-}) (q - q_{-}), \notag
\end{align}
where $N = d \phi / d q$.}

\subsection{Interior-Point Method} \label{ipm}
\diff{Classically, LCPs are solved using active-set methods \cite{dirkse1995path}, which strictly enforce complementarity at each iteration. An alternative approach is interior-point methods \cite{kojima1991unified, nocedal2006numerical}, which relax these conditions during intermediate iterations, only satisfying these constraints in the limit.}

LCPs can be generally formulated as:
\begin{equation}
	\begin{array}{ll}
	\mbox{find}       & x, y, z \\
	\mbox{subject to} & E x + F y + f = 0,\\
					  & G x + H y + z + h = 0,\\
					  & y \circ z = 0,\\
					  & y, z \geq 0,\\
\end{array} \label{lcp_problem}
\end{equation}
with decision variables $x \in \mathbf{R}^n$, $y, z \in \mathbf{R}^m$ and problem data $\theta = (E, F, G, H, f, h)  \in \mathbf{R}^{n \times n} \times \mathbf{R}^{n \times m} \times \mathbf{R}^{m \times n} \times \mathbf{R}^{m \times m} \times \mathbf{R}^n \times \mathbf{R}^m = \mathbf{R}^p$. Interior-point methods parameterize (\ref{lcp_problem}) by a  central-path parameter $\kappa \in \mathbf{R}_{+}$ that relaxes the following bilinear constraint:
\begin{equation}
    y \circ z = \kappa \textbf{1},
\end{equation}
where $\textbf{1}$ a vector of ones. 

The equality and relaxed bilinear constraints form a residual vector or solution map, $r: \mathbf{R}^{n + 2m} \times \mathbf{R}^p  \times \mathbf{R}_{+} \rightarrow \mathbf{R}^{n + 2m}$, that takes $w = (x, y, z) \in \mathbf{R}^{n + 2m}$, the problem data, and central-path parameter as inputs. The problem data and central-path parameter are fixed during optimization. In the context of contact dynamics, these data encode the mechanical properties of the robots, its current configuration and velocity, and properties of the environment like friction coefficients.
Newton or quasi-Newton methods are used to find search directions that reduce the norm of the residual and a backtracking line search is employed to ensure that the inequality constraints are strictly satisfied for candidate points at each iteration. Once the residual is optimized to a desired tolerance, the central-path parameter is decreased and the new subproblem is warm-started with the current solution and then optimized. This procedure is repeated in order to find solutions to (\ref{lcp_problem}) with $\kappa \rightarrow 0$ until the central-path parameter, also referred to as complementary slackness, is below a desired tolerance.

\diff{Importantly, our interior-point method utilizes a predictor-corrector algorithm \cite{mehrotra1992implementation} that leads to significantly improved convergence. First, the corrector step modifies the pure Newton search direction and typically reduces the number of iterations required for convergence by half (compared to the pure search direction). Second, the central-path parameter is adapted at each iteration to prevent premature numerical ill-conditioning. In practice, we find that this approach is significantly more reliable and has improved convergence behavior compared to prior work that employed relaxed complementarity conditions \cite{manchester2020variational}.}

\subsection{Implicit Differentiation} \label{ift}
\diff{In addition to simulating contact by solving a feasibility problem, we would like to compute gradients of these dynamics, requiring us to differentiate through an optimization problem. This is accomplished with implicit differentiation \cite{dini1907lezioni}.}

An implicit function, $r : \mathbf{R}^k \times \mathbf{R}^p \rightarrow \mathbf{R}^k$, is defined such that:
\begin{equation}
	r(w^*; \theta) = 0, \label{implicit_function}
\end{equation}
for solutions $w^* \in \mathbf{R}^k$ and problem data $\theta \in \mathbf{R}^p$. At a stationary point, $w^*(\theta)$, the sensitivity of the solution with respect to the problem data, i.e., $\frac{\partial z}{\partial \theta}$, can be computed by utilizing the implicit-function theorem \cite{dini1907lezioni}. We expand (\ref{implicit_function}) to first order:
\begin{equation}
	\frac{\partial r}{\partial w} \delta w + \frac{\partial r}{\partial \theta} \delta \theta = 0,
\end{equation}
and then solve for $\delta w$: 
\begin{equation}
	\frac{\partial w^*}{\partial \theta} = -\Big(\frac{\partial r}{\partial w}\Big)^{-1} \frac{\partial r}{\partial \theta}, \label{solution_sensitivity}
\end{equation}
to compute the sensitivities.

\diff{For the interior-point method \eqref{lcp_problem}, the residual is:
\begin{equation}
    r(w; \theta) = \begin{bmatrix} E x + F y + f \\ G x + H y + z + h \\ y \circ z - \kappa \textbf{1}\end{bmatrix}.
\end{equation}
A differentiable interior-point method is summarized in Algorithm \ref{ip_algo}. Importantly, we can compute gradients for intermediate results, corresponding to non-zero values for the central-path parameter, i.e., $\kappa_{\mbox{grad}} \neq 0$. For additional details about differentiating through intermediate results of contact dynamics that are solved with interior-point methods, see \cite{howell2022dojo}.}

\begin{algorithm}[t]
    \caption{Differentiable Interior-Point Method}\label{ip_algo}
    \begin{algorithmic}[1]
    \Procedure{Optimize}{$x, \theta$}
    \State \textbf{Settings}: $\beta = 0.5, \gamma = 0.1, \epsilon_{\kappa} = 10^{-6}, \epsilon_{r} = 10^{-8}$
    \State \textbf{Initialize}: $y, z = \mathbf{1}, \kappa = 0.1, \kappa_{\mbox{grad}} = 10^{-4}$
    \State \textbf{Until} $\kappa < \epsilon_{\kappa}$ \textbf{do} 
    \State \indent $\Delta w = (\frac{\partial r}{\partial w})^{-1} r(w; \theta, \kappa)$
    \State \indent $\alpha \leftarrow 1$
    \State \indent \textbf{Until} $(y, z) - \alpha (\Delta y, \Delta z) > 0$ \textbf{do} $\alpha \leftarrow \beta \alpha$
    \State \indent \textbf{Until} $\|r(w-\alpha\Delta w; \theta, \kappa)\| < \|r(w; \theta, \kappa)\|$ \textbf{do}
    \State \indent \indent $\alpha \leftarrow \beta \alpha$
    \State \indent $w \leftarrow w - \alpha \Delta w$
    \State \indent \textbf{If} $\|r(w; \theta, \kappa)\| < \epsilon_{r}$ \textbf{do} $\kappa \leftarrow \gamma \kappa$
    \State $\frac{\partial w}{\partial \theta} \leftarrow \textbf{Differentiate}(w, \theta, \kappa_{\mbox{grad}})$ \Comment{Eq. \ref{solution_sensitivity}}
    \State \textbf{Return} $w, \frac{\partial w}{\partial \theta}$ 
    \EndProcedure
    \end{algorithmic}
\end{algorithm}

\section{Contact-Implicit Model Predictive Control} \label{cimpc}
\diff{In this section we present Contact-Implicit Model Predictive Control, a tracking policy for systems that make and break contact with their environments. First, we formulate time-varying LCP contact dynamics that are selectively approximated about a reference trajectory. Then, we devise a fast solver for the resulting LCP. Next, we discuss how to compute smooth gradients through the dynamics. A bi-level planning formulation, which utilizes these dynamics for direct trajectory optimization \cite{von1992direct}, follows. To enable the policy to work well in environments with uncertain terrain we propose a contact-height heuristic. Finally, we summarize the approach and provide an algorithm for CI-MPC.}

\subsection{Time-Varying Contact Dynamics} \label{con_dyn}
\diff{We formulate alternative LCP dynamics that utilize a reference trajectory, resulting in the following mappings: 
\begin{align}
    &M_{+} \leftarrow M(\bar{q}), 
    &&M_{-} \leftarrow M(\bar{q}_{-}), \notag \\ 
    &C \leftarrow C(\bar{q}, (\bar{q} - \bar{q}_{-}) / h), 
    &&J \leftarrow J(\bar{q}),  \\ 
    &B \leftarrow B(\bar{q}), 
    && P \leftarrow P(\bar{q}), \notag \\
    & \phi \leftarrow \phi(\bar{q}) + N(\bar{q}) (q - \bar{q}). \notag
\end{align}
The LCP problem, formulated for an interior-point method, has the form:
\begin{align}
	\begin{array}{ll}
	{\mbox{find}} & w \\
	\mbox{subject to} 
	& C (w - \bar{w}) + D (\theta - \bar{\theta}) = 0 \\
	& \gamma \circ s_{\phi}  = \kappa \textbf{1}, \\
	& \psi \circ s_{\psi} = \kappa \textbf{1}, \\
	& \beta \circ \eta = \kappa \textbf{1}, \\
	& \gamma, \psi, \beta, \eta, s_{\phi}, s_{\psi} \geq 0, \\
    \end{array} \label{linearized_feas_prob}
\end{align}
with decision variables $w = (q, \gamma, \beta, \psi, \eta, \beta, s_{\phi}, s_{\psi})$, where slack variables, $s_{\phi}, s_{\psi} \in \mathbf{R}$, are introduced for convenience.}

\diff{Importantly, $C$ and $D$ are matrices that define a linear system of equations resulting from approximations about the reference trajectory and they are pre-computed offline. These contact dynamics:
\begin{equation}
    q_{t+1} = \textbf{LCP}_t(q_{t-1}, q_t, u_t),
\end{equation}
$\textbf{LCP}_t : \mathbf{R}^{n_q} \times \mathbf{R}^{n_q} \times \mathbf{R}^m \rightarrow \mathbf{R}^{n_q}$, solve (\ref{linearized_feas_prob}) and return the configuration at the next time step. The contact forces at the current time step can also be returned.}

\subsection{Fast Contact Dynamics}
The most expensive procedure in evaluating the LCP and computing gradients of a solution is solving the linear system of equations:
\begin{equation}
     R_w \Delta w = r, \label{eq:naive_solve}
\end{equation}
\diff{required by the interior-point method, where $R_w = \partial r / \partial w$, and $\Delta w$ is the new search direction.}

To reduce the computational cost of this routine, we exploit both the sparsity pattern and the property that most of $R_w$ remains constant across iterations and, therefore, can be pre-factorized offline \cite{yamazaki2017structure}. 

\diff{We partition the LCP variables \eqref{lcp_problem} as follows: $x = q$, $y = (\gamma, \psi, \beta)$, and $z = (\eta, s_{\phi}, s_{\psi})$, and similarly split the residual: $r = (r_x, r_y, r_z)$. The Jacobian's sparsity pattern is:}
\begin{align}
R_w &= \begin{bmatrix}
	    E & F & 0 \\
	    G & H & I \\
	    0 & \mbox{\textbf{diag}}(z) & \mbox{\textbf{diag}}(y)
	    \end{bmatrix}, \label{Rw_sparsity}
\end{align}
where $I$ denotes the identity matrix. By exploiting sparsity in the third row of (\ref{Rw_sparsity}), we can form the following condensed system:
\begin{align}
    \begin{bmatrix}
        E & F \\
        G & \tilde{H} \\
    \end{bmatrix}
    \begin{bmatrix}
        \Delta x \\
        \Delta y \\
    \end{bmatrix}
    = 
    \begin{bmatrix}
        r_x \\
        \tilde{r}_y \\
    \end{bmatrix} \Leftrightarrow \tilde{R}_w \: \Delta \tilde{w} = \tilde{r},
    \label{Rw_reduced}
\end{align}
where:
\begin{align}
    \tilde{H} &= H - \mbox{\textbf{diag}}(y^{-1} \circ z), \\
    \tilde{r}_y &= r_y - y^{-1} \circ r_z, \\
    \Delta z &= y^{-1} \circ (r_z - z \circ \Delta y),
    \label{eq:delta_w3}
\end{align}
and $y^{-1}$ denotes the element-wise reciprocal of vector $y$. This term is always well-defined because a line search enforces $y > 0$ at each iteration. 

To solve for $\Delta \tilde{w}$, we leverage the fact that, apart from the bottom-right block, $\tilde{R}_w$ can be computed offline. We perform a QR decomposition on the Schur complement of (\ref{Rw_reduced}): 
\begin{align}
    Q, R \leftarrow \mbox{\textbf{QR}}(\tilde{H} - G E^{-1} F),
\end{align}
and then solve for the search directions:
\begin{align}
    \Delta y &= - R^{-1} Q^T (G E^{-1} r_x - \tilde{r}_y), \\
    \Delta x & = E^{-1} (r_x - F \Delta y).
\end{align}
Additionally, $E^{-1}$, $G E^{-1}$, and $G E^{-1} F$ are precomputed offline. Finally, after solving for $\Delta \tilde{w}$, we obtain $\Delta z$ with cheap vector-vector operations (\ref{eq:delta_w3}).

For a system with configuration dimension $n_q$ and $c$ contact points, the computational complexity of solving (\ref{eq:naive_solve}) with a naive approach is $O\Big((n_q+2cd)^3 \Big)$. Our structure-exploiting approach is $O(8c^3d^3)$ during the online phase. \diff{In practice, this provides a factor of 15 speed-up, compared to LAPACK LU, for evaluating the LCP dynamics across all the robotic systems presented in this paper and, in turn, results in a factor of 2.5 speed-up for CI-MPC.}

\begin{algorithm}[t]
    \diff{
	\caption{Contact-Implicit Model Predictive Control}\label{alg:mpc}
	\begin{algorithmic}[1]
		\Procedure{Policy}{}
		\State \textbf{Offline}  
		\State \indent $\bar{\tau} \leftarrow$ generate reference trajectory
		\State \indent $\mbox{\textbf{LCP}}_t \leftarrow$ generate fast contact dynamics
		\State \indent $H \leftarrow$ set planning horizon, 
		\State \textbf{Online}  
		\State \indent \textbf{For} $i = 1, \ldots, \infty$
		\State \indent \indent $u \leftarrow \pi(x)$ \Comment{Eq. \eqref{ci_mpc_policy}}
		\State \indent \indent $x \leftarrow \textbf{dynamics}(x, u)$
		\State \indent \textbf{End} 
		\EndProcedure{\textbf{End}}
	\end{algorithmic}
    }
\end{algorithm}

\subsection{Gradients}\label{ipm_diff}
\diff{Contact dynamics gradients are computed by differentiating through the LCP problem with respect to data: $\theta = (q_{--}, q_{-}, u)$, which could also include the time step, friction coefficients, and other system values like masses or inertia terms.}

\diff{At fixed points, $w^*_{\kappa}$, parameterized by a (potentially non-zero) central-path value, gradients are computed using implicit-differentiation. Importantly, at solution where $\kappa \approx 0$, this approach will return subgradients, which often fail to provide useful information through contact events. However, we exploit intermediate results from the interior-point solver, where $\kappa \neq 0$, in order to compute smooth gradients. Large values of $\kappa$ will produce smoother gradients than those computed with small values of $\kappa$, which more closely approximates a true subgradient at nondifferentiable points. Practically, we find that smooth gradients, computed using intermediate results, provides information through contact events.}

\begin{figure}[t]
	\centering
		\includegraphics[width=.45\textwidth]{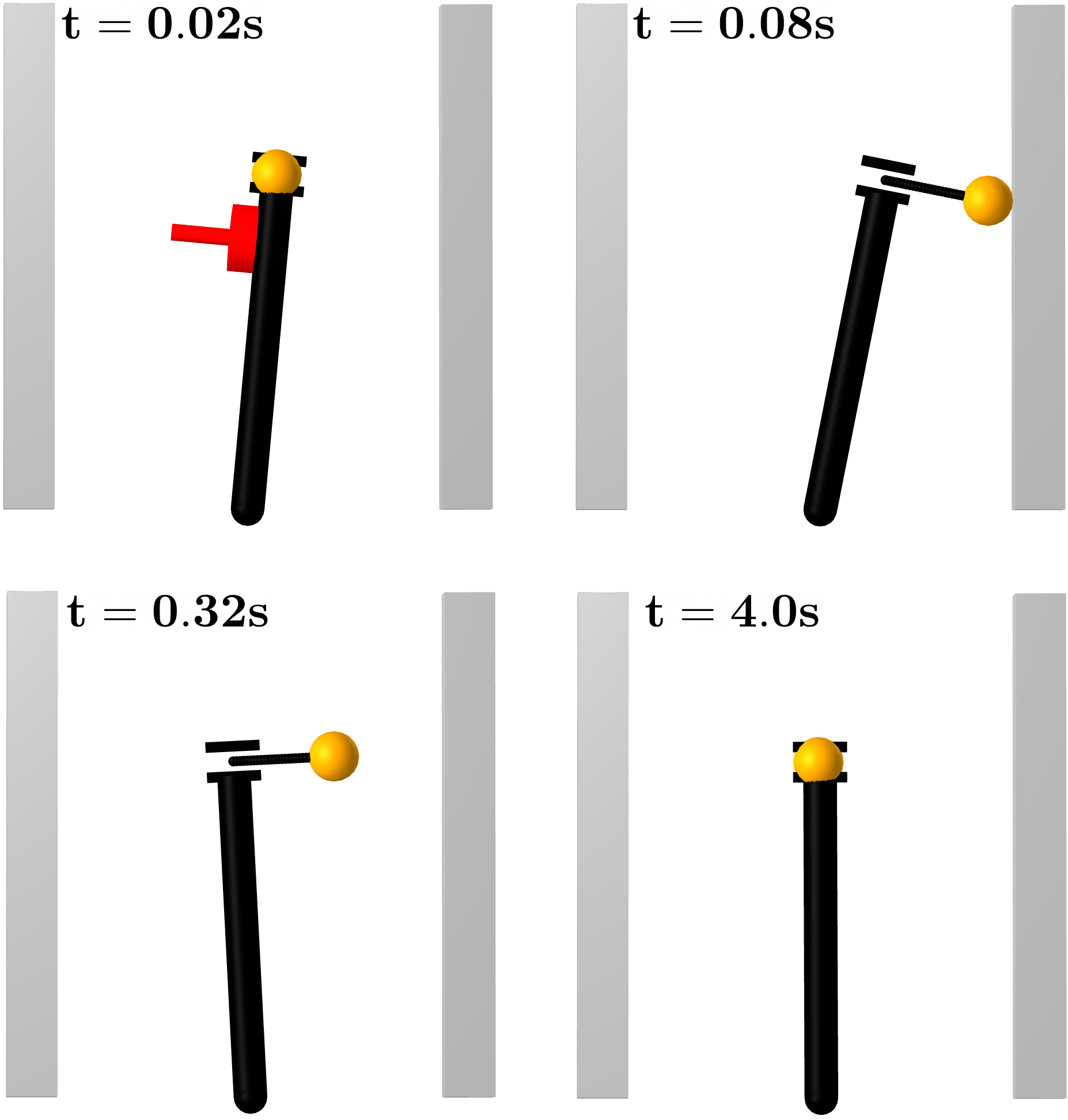}
		\caption{PushBot performing push recovery. A disturbance (red) creates an impulse on the system and the policy generates a new contact sequence that extends the prismatic joint toward the right wall in order to make contact. After stabilizing, PushBot pushes against the wall, eventually breaking contact, in order to return to the nominal upright configuration.}
	\label{pushbot_push_recovery}
\end{figure}

\subsection{Planning}
\diff{
The CI-MPC policy is formulated as: 
	\begin{equation}
	u = \pi(x)
	\begin{cases}
		\begin{array}{ll}
			\underset{x_{1:H}, u_{1:H}}{\mbox{minimize }}  & \sum \limits_{t = 1}^{H} \frac{1}{2}(x_t - \bar{x}_t)^T Q_t (x_t - \bar{x}_t) \label{ci_mpc_policy} \\
			&  + \frac{1}{2}(u_t - \bar{u}_t)^T R_t (u_t - \bar{u}_t)\\
			\mbox{subject to } & x_{t+1} =\textbf{LCP}_t(x_t, u_t), \\
			& \quad t = 1, \dots, H-1, \\
			& x_1 = x,
		\end{array}
	\end{cases}
\end{equation}
comprising an upper-level planning problem (\ref{trajectory_optimization}) that optimizes a trajectory: $\tau = (x_1, u_1,\dots, x_H, u_H) \in \mathbf{R}^{(2 n_q + m) H}$ of configurations and controls over a horizon $H$ using a state representation: $x_t = (q^{(t)}_{t-1}, q^{(t)}_{t})$, with two configurations. This problem is solved using a direct trajectory optimization approach. Lower-level LCP problems (\ref{feas_prob}) enforce the dynamics with the first state $x_1$ fixed. For convenience, we overload notation for the LCP dynamics:
\begin{align}
	\textbf{LCP}_t(x_t, u_t) &= \begin{bmatrix} q^{(t)}_{t} \\  \textbf{LCP}_t(q_{t-1}^{(t)}, q_t^{(t)}, u_t) \end{bmatrix}, 
\end{align}
for state-based LCP dynamics, and define constraints, $k_t(x_t, u_t, x_{t+1}) = x_{t+1} - \textbf{LCP}_t(x_t, u_t)$, that couple states across adjacent time steps. The constraint Jacobian:
\begin{align}
	\nabla k = \begin{bmatrix} 
		-B_1 & I & 0 & 0 & 0 & 0 & \phantom{\cdots}\\
		0 & -A_2 & -B_2 & I & 0 & 0 & \phantom{\cdots}\\
		0 & 0 & 0 & -A_3 & -B_3 & I & \phantom{\cdots}\\ 
		  &   &   &      &      &   & \ddots
	\end{bmatrix},
\end{align}
where $k = (k_1, \cdots, k_{H-1}) \in \mathbf{R}^{2n_q(H-1)}$, is comprised of one-step dynamics Jacobians:
\begin{equation}
	A_t = \left[\begin{array}{c c} 0 & I \\ \frac{\partial \textbf{LCP}_t}{\partial q_{t-1}} & \frac{\partial \textbf{LCP}_t}{\partial q_{t}} \end{array} \right],\, B_t = \begin{bmatrix} 0 \\ \frac{\partial \textbf{LCP}_t}{\partial u_t} \end{bmatrix}.
\end{equation}
}

\begin{table}[t]
	\centering
    \caption{Comparison between CI-MPC and MIQP policies for PushBot example. For a fixed replanning rate of 25 Hz, we report the mean and standard deviations for the optimization times and compare this to the associated time budget (0.04 s). Both policies successfully regulate the system around the equilibrium point. However, the MIQP policy is slower than real-time, whereas the CI-MPC policy always remains within time budget, ensuring real-time performance.}

	\begin{tabular}{c c c}
		\toprule
	    \textbf{Policy} &
		\textbf{Planning Time} &
		\textbf{Real-Time} \\
		\toprule
		CI-MPC & \boldmath $0.014 \pm 0.027 \mbox{\textbf{s}}$ \unboldmath & \checkmark \\ 
		MIQP   & $0.18  \pm 0.09$ s & \ding{53} \\ 
		\toprule
	\end{tabular}
    \label{fig:pushbot_results}
\end{table}

\diff{The planning objective is a convex quadratic function \eqref{quadratic_objective} and velocities are penalized using finite-difference approximations. }Because the problem is lifted by using states comprising two configurations, these costs do not introduce coupling across more than one time step. The resulting Hessian of the objective:
\begin{equation}
	\nabla^2 J = \begin{bmatrix} R_1 & 0 & 0 & 0 \\ 0 & Q_2 & 0 & 0 \\ 0 & 0 & R_2 & 0 \\ 0 & 0 & 0 & \ddots \end{bmatrix},
\end{equation}
and its inverse:
\begin{equation}
	(\nabla^2 J)^{-1} = \begin{bmatrix} R^{-1}_1 & 0 & 0 & 0 \\ 0 & Q^{-1}_2 & 0 & 0 \\ 0 & 0 & R^{-1}_2 & 0 \\ 0 & 0 & 0 & \ddots \end{bmatrix},
\end{equation}
are block diagonal and are pre-computed offline.

The resulting KKT system:
\begin{equation}
	\begin{bmatrix} \nabla^2 J & \nabla k^T \\ \nabla k & 0 \end{bmatrix} \begin{bmatrix} \Delta \tau \\ \Delta \nu \end{bmatrix} = \begin{bmatrix} \nabla J + \nabla k^T \nu \\ k \end{bmatrix}, \label{trajopt_kkt}
\end{equation}
\diff{with dual variables $\nu \in \mathbf{R}^{2 n_q (H-1)}$ associated with the constraints, uses a Gauss-Newton approximation of the constraints when computing the Hessian of the Lagrangian and is solved using a sparse $\mbox{LDL}^T$ solver. In the following experiments we utilize QDLDL, a general-purpose sparse solver, for its efficient implementation \cite{davis2005algorithm}.}

\subsection{Contact-Height Heuristic} \label{sec:heuristic}
To enable the policy to robustly adapt to unknown variations in terrain height, we employ a simple heuristic that we find to be effective in practice. The policy maintains a height estimate, $a \in \mathbf{R}^c$, for each contact and utilizes a modified signed-distance function:
\begin{equation}
    \phi_{MPC}(q) = \phi(q) + a,
\end{equation}
that is updated using the current contact height. When contact is detected, the height estimate is updated. In simulation, a threshold on the impact-force magnitude is set; and in practice, force sensors can reliably detect such an event.

This simple heuristic does not affect the structure of (\ref{Rw_sparsity}) and only requires $c$ more addition operations to compute $r$ when evaluating the fast contact dynamics (\ref{eq:naive_solve}). In our experiments, we find the heuristic to be effective and reliable across unknown terrain for the systems tested.

\begin{figure}[t]
	\begin{center}
		\includegraphics[width=0.25\textwidth]{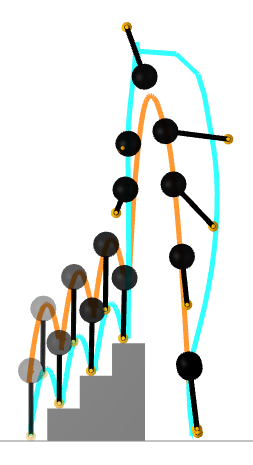}
	\end{center}
	\caption{Hopper in 2D performing parkour. The system tracks the body (orange) and foot (blue) reference trajectories while ascending three stairs before performing a front flip.}
	\label{hopper_parkour}
\end{figure}

\subsection{Algorithm}
\diff{CI-MPC comprises offline and online stages. The offline stage generates a reference trajectory along with a set of time-varying LCP problems. In this work we employ contact-implicit trajectory optimization \cite{manchester2020variational} to design these references. Additionally, a planning horizon, typically less than the total behavior duration, is specified. During the online stage, planning is performed (\ref{trajectory_optimization}) for the current state over the specified horizon, and the optimized control trajectory is used to compute an control that is applied to the system. The CI-MPC policy is summarized in Algorithm \ref{alg:mpc}.}

\subsection{Heuristics}
\diff{To enable real-time performance for the policy \eqref{ci_mpc_policy}, a number of heuristics are employed. First, the planning problem is only solved approximately. Instead of optimizing until convergence, a fix number of iterations are performed and then the current best solution is returned. This enables the current plan to be improved but significantly reduces the total computation required. Generally, we find that performance is greatly improved by returning approximate solutions quickly, enabling replanning with newer state information, compared to returning higher quality solutions to planning problems that are utilizing older state information. Second, the policy extensively utilizes warm starting. Providing the optimizer with a good initial guess for the solution greatly reduces the number of iterations required to converge. Initially, the policy utilizes the reference trajectory. At subsequent evaluations, the previous best solution is used.
Third, the LCP problems are solved for a single value of the central path parameter instead of a sequence that converges to zero. In practice, we find that $\kappa \approx 1\mbox{e-}4$ is a good balance between computation time, physical accuracy, and gradient smoothness. Note, when verifying the performance of the policy in simulation, we solve the nonlinear contact dynamics complementarity problem (\ref{feas_prob}) to convergence, i.e., $\kappa = 1\mbox{e-}6$.}

\section{Results} \label{results}

We demonstrate the CI-MPC algorithm in simulation and on hardware by controlling a variety of robotic systems that make and break contact with their environments. In the examples we show that the policy can generate new contact sequences online; is robust to disturbances, model mismatch, and unknown terrain; and is faster than real-time, see Table \ref{fig:timing_table}. The code, including a Julia implementation of the policy and all of the experiments, is available at: 
\begin{center}
\url{https://github.com/dojo-sim/ContactImplicitMPC.jl}.
\end{center}
\begin{table}[t]
	\centering
    \caption{Comparison between CI-MPC and the Raibert heuristic for a hopper system on 4 scenarios: flat, sinusoidal, and piecewise linear terrains; and a parkour stunt (Fig. \ref{hopper_parkour}). For each terrain profile, we report the number of hops achieved by the policy. For the parkour scenario, we report if the stunt is successfully completed.}
    
	\begin{tabular}{c c c c c}
		\toprule
	    \textbf{Policy} &
		\textbf{Flat} &
		\textbf{Sinusoidal} &
		\textbf{Piecewise} &
		\textbf{Parkour}\\
		\toprule
		CI-MPC & $+100$ & $+100$ & $+100$ & \checkmark \\ 
		Raibert & $+100$ & $+100$ & $+100$ & \ding{53} \\ 
		\toprule
	\end{tabular}
    \label{fig:hopper_results}
\end{table}

\subsection{Simulation}
We verify the policy performance in simulation  where we solve the nonlinear contact dynamics complementarity problem (\ref{feas_prob}) to convergence. Additionally, all examples are simulated using a different sample rate, typically 5-10$\times$ faster than the reference trajectory, in order to ensure that the policy is robust to sampling rates. \diff{For example, if the reference is optimized with a time step of $0.1$ seconds, then we simulate the nonlinear dynamics with a smaller time step, $0.01$ or $0.02$ seconds.}

\emph{PushBot:}
In this example, we demonstrate that our policy can generate qualitatively new, unspecified contact sequences online in order to respond to unplanned disturbances. The system, PushBot, is modeled as an inverted pendulum with a prismatic joint located at the end of the pendulum (Fig. \ref{pushbot_push_recovery}). There are two control inputs: a torque at the revolute joint and a force at the prismatic joint. The system is located between two walls and has two contact points, one between the prismatic-joint end effector and each wall.

PushBot is tasked with remaining vertical and the policy utilizes a reference trajectory that does not include any contacts. When we apply a large impulse to the system, the policy generates a behavior that commands the prismatic joint to push against the wall in order to stabilize. By tuning the policy's cost function we can generate different behaviors, including maintaining contact to stabilize and pushing against the wall in order to return to the nominal position. The latter behavior is shown in Fig. \ref{pushbot_push_recovery}.

\begin{figure}[t]
	\centering
		\includegraphics[width=.48\textwidth]{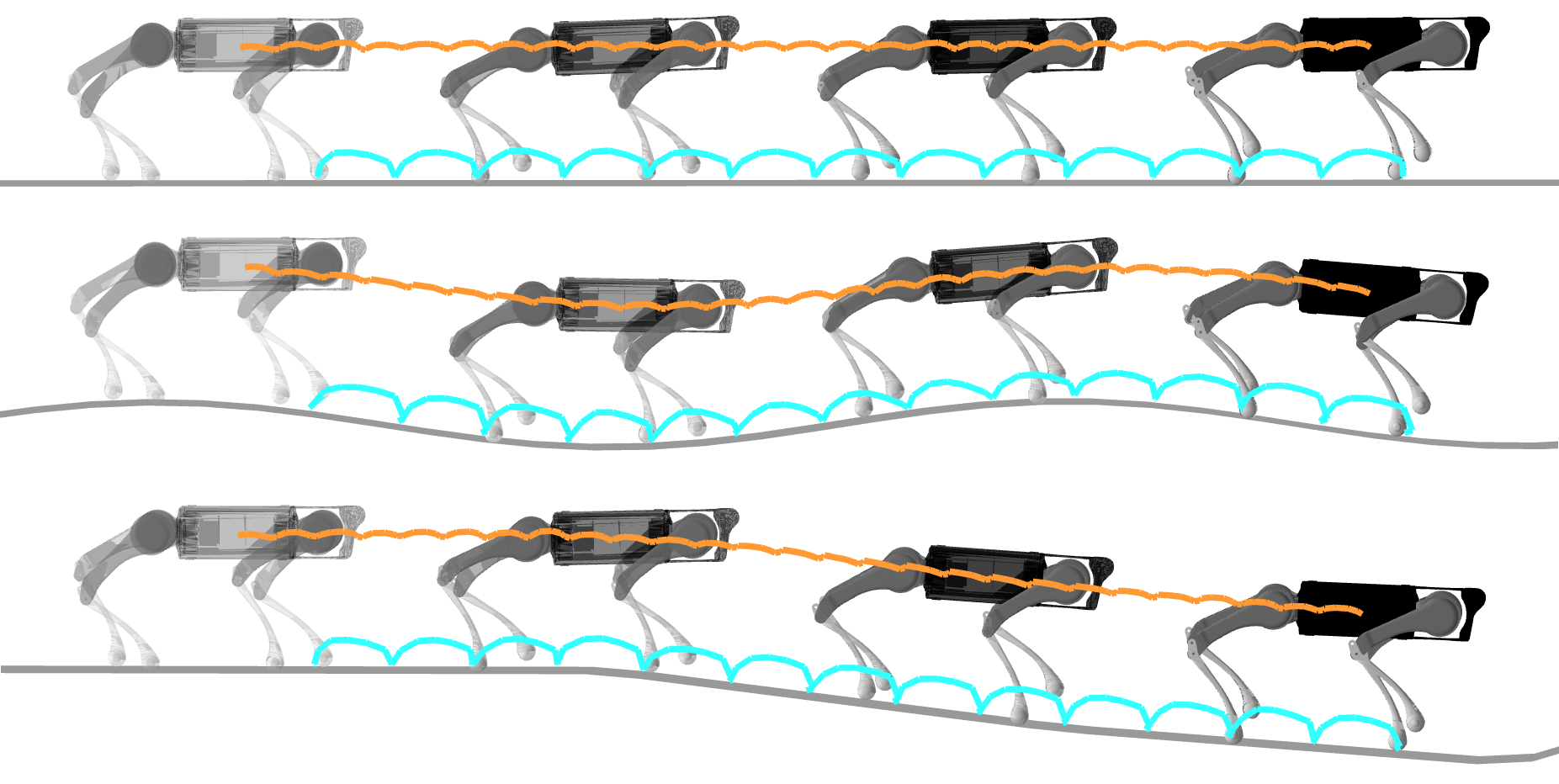}
		\caption{Planar quadruped walking over uneven terrain. The reference gait is optimized for flat ground.  Our CI-MPC policy, with orange center-of-mass and blue foot position trajectories, is able to adapt online to the unmodeled variation in terrain and track the reference trajectory.}
	\label{quadruped_traj}
\end{figure}

We compare CI-MPC to a method that relies on a mixed-integer quadratic progam (MIQP) formulation \cite{bemporad1999control} applied to a simplified version of the PushBot (an inverted pendulum between two stiff walls). The MIQP minimizes a quadratic objective function subject to piecewise-linearized dynamics. Each linear dynamics domain corresponds to a single contact mode, and discrete decision variables are introduced to encode contact mode switches. Our CI-MPC approach is fast enough to be run online, however, this is not the case for the MIQP policy, as shown in Table \ref{fig:pushbot_results}. Moreover, the complexity of the MIQP increases exponentially with the number of contact modes, making it an intractable approach for more complex systems. \diff{Warm-starting the MIQP \cite{marcucci2020warm} may make this approach more amenable to online optimization.}

\emph{Hopper:}
Inspired by the Raibert Hopper \cite{raibert1989dynamically}, we model a 2D hopping robot with $n_q = 4$ generalized coordinates: lateral and vertical positions, body orientation, and leg length, respectively; $m = 2$ controls: body moment, e.g., controlled with an internal reaction wheel, and leg force; and a single contact at the foot. 

The centroidal-dynamics modeling assumption we make---consistent with Raibert's work---is to locate the leg and foot mass at the body's center of mass. This results in a configuration-independent mass matrix and no bias term in the dynamics. 

\begin{figure}[t]
		\centering
		\includegraphics[width=.45\textwidth]{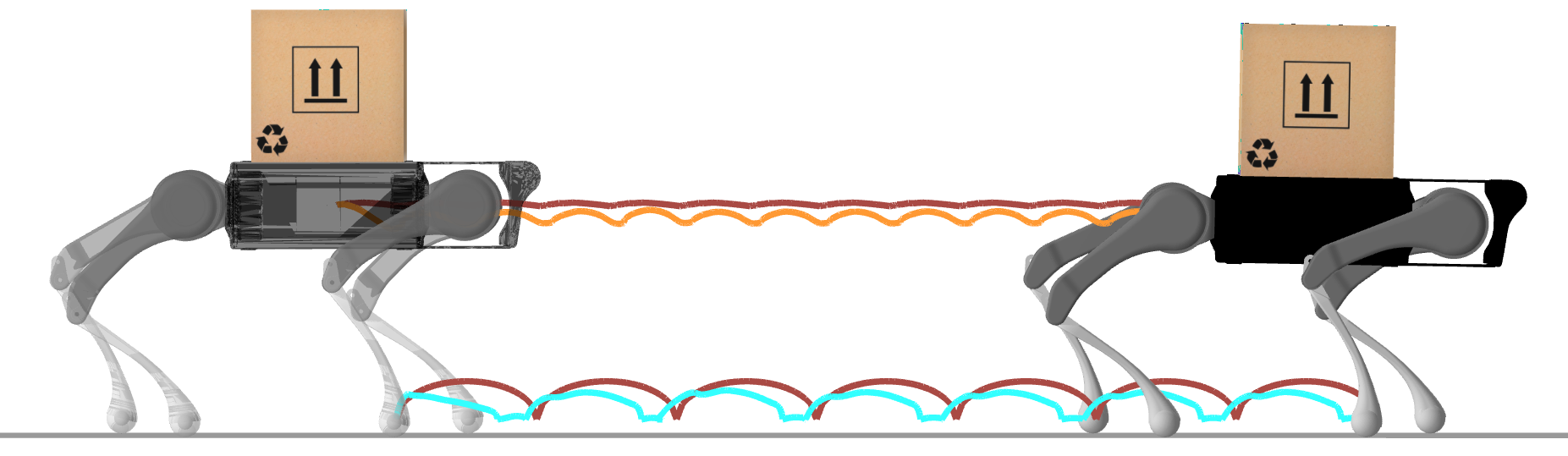}
		\caption{Quadruped tracking a reference trajectory (red) while carrying an unmodeled $3$-kg payload. We depict the torso (orange) and front-left foot (blue) trajectories.}
	    \label{quadruped_payload}
\end{figure}

The hopper is tasked with locomoting over unknown terrain. The CI-MPC policy uses a reference trajectory that is optimized with a flat surface and no incline. We compare our policy to the Raibert heuristic, which we similarly tune for flat ground and no incline. We observe that our policy is able to adapt to the varying surface heights that range from $0\mbox{-}24\:$cm and that the robot can slip multiple times and is able to recover while traversing steep inclines. We find that, when tuned well, the Raibert heuristic also works very well on terrains

Additionally, we task the hopper with climbing a staircase and executing a front flip (Fig. \ref{hopper_parkour}). This complex trajectory cannot be directly executed using the Raibert heuristic as it is not a periodic hopping gait. Our policy, however, successfully tracks this complex trajectory, illustrating the more general capabilities of CI-MPC. Results are summarized in Table \ref{fig:hopper_results}.  

\emph{Planar quadruped:}
We model a planar quadruped with $n_q = 11$ configuration variables and $m = 8$ control inputs. The system has four contacts, one at each point foot. 

The quadruped is tasked with moving to the right over three different terrains: flat, sinusoidal, and piecewise-linear surfaces (Fig. \ref{quadruped_traj}). Additionally, we test the robustness of the CI-MPC policy by introducing model mismatch. We provide the policy with the nominal model of the quadruped while the simulator uses a quadruped with a $3$-kg payload, representing $25\%$ of its nominal mass. Despite the unmodeled load, the policy successfully tracks the nominal gait with good performance. 

We note that the same CI-MPC policy was used across all quadruped experiments and no retuning was required to transfer from the nominal case (flat terrain, no payload) to more complex scenarios. Further, it is easy and intuitive to rapidly retune the policy in order to achieve improved tracking performance in the 
other scenarios.

\emph{Planar biped:}
We model a planar biped based on Pratt's Spring Flamingo \cite{pratt2000thesis} with $n_q = 9$ configuration variables and $m = 7$ control inputs. The system is modeled with four contact points, one at the toe and heel of each foot.

\begin{figure}[t]
		\centering
		\includegraphics[width=.45\textwidth]{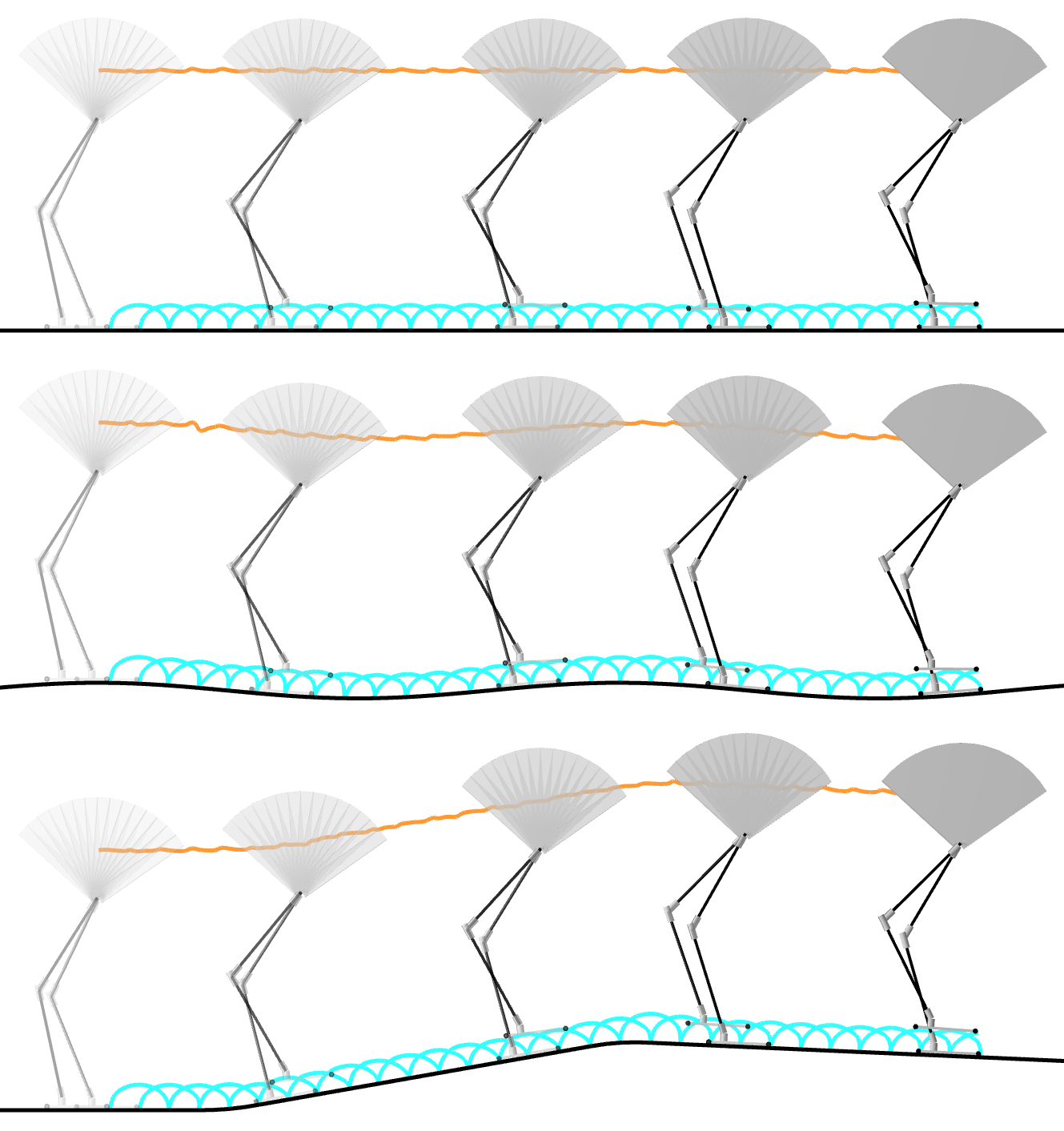}
		\caption{Biped walking from left to right across flat (top), sinusoidal (middle), and piecewise linear (bottom) terrain using the same policy.}
	    \label{biped_tracking}
\end{figure}

The biped is tasked with moving to the right over three different terrains: flat, sinusoidal, and piecewise-linear surfaces (Fig. \ref{biped_tracking}) using the same policy. In Table \ref{fig:biped_results}, we compare this to Pratt's policy \cite{pratt2000thesis}, which relies on a state-machine architecture and a number of proportional-derivative controllers. Our CI-MPC policy---with no additional tuning---can easily walk on all of the terrains and reliably walks up inclines of up to ten degrees. Pratt reports that Spring Flamingo can only walk up inclines of five degrees without requiring the controllers to be re-tuned \cite{pratt1998intuitive}.

\begin{figure}[t]
	\begin{center}
		\includegraphics[width=0.45\textwidth]{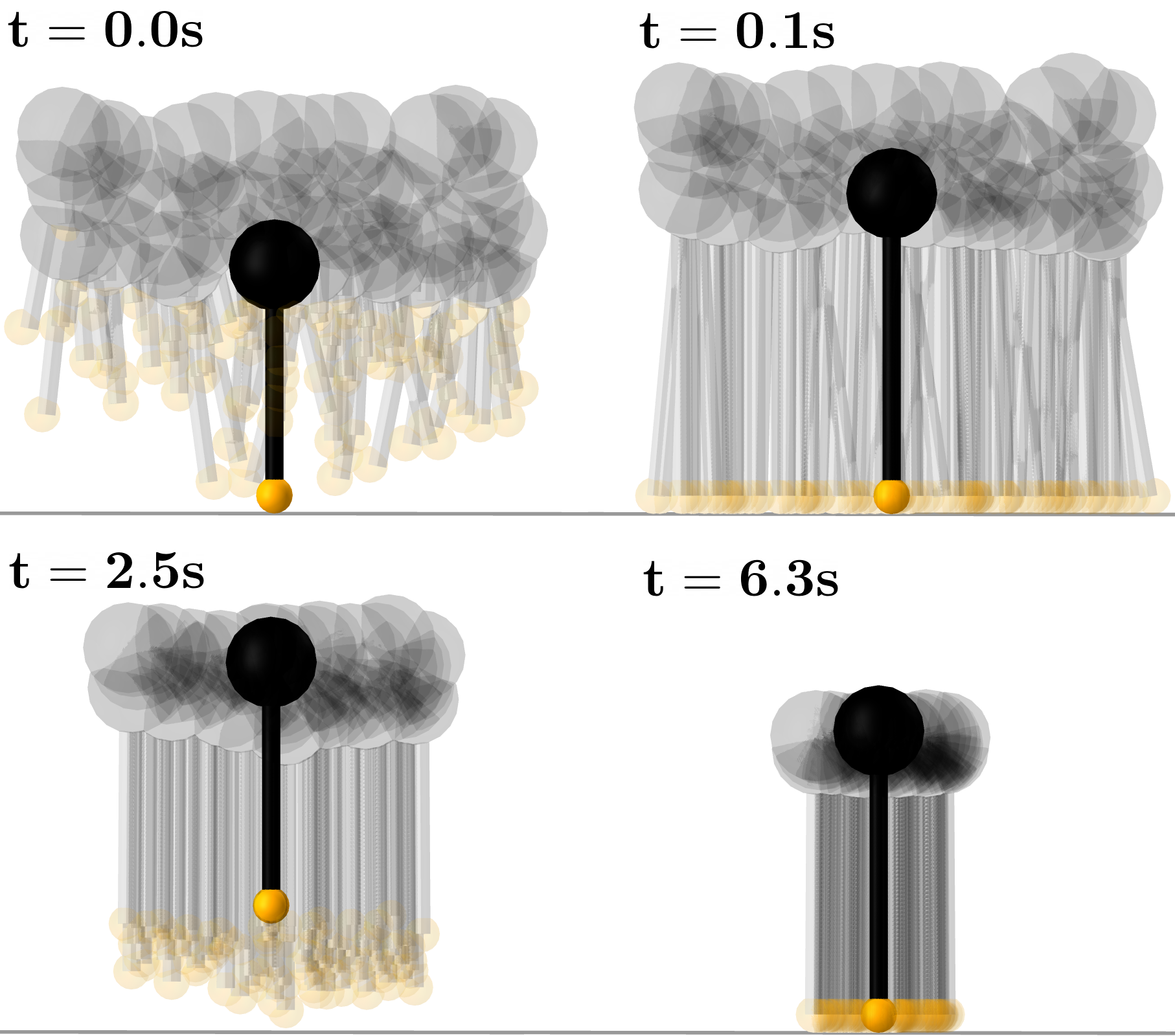}
	\end{center}
	\begin{center}
		\includegraphics[width=0.45\textwidth]{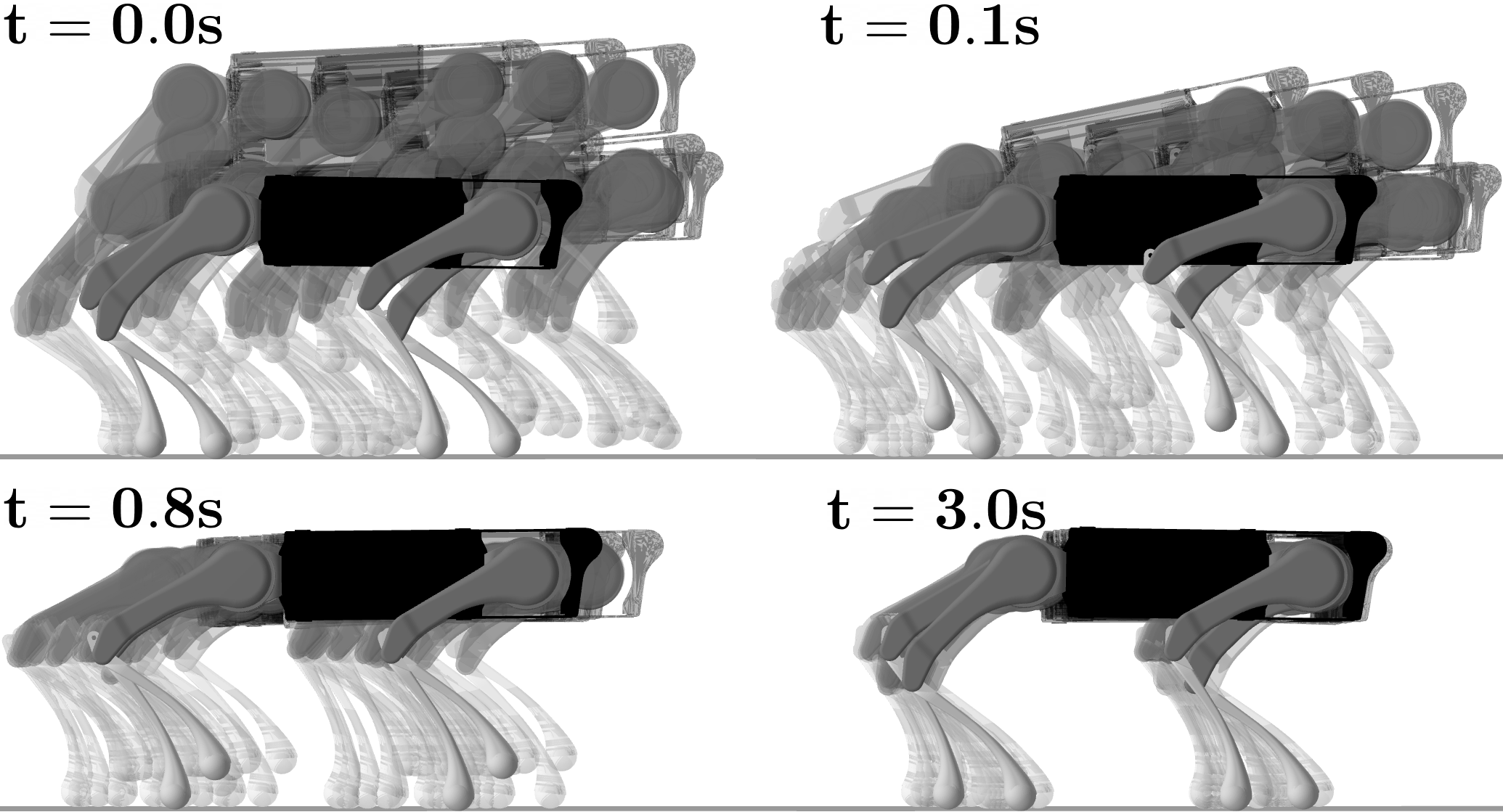}
	\end{center}
	\caption{Monte Carlo simulations of initial conditions for systems tracking a reference trajectory. 100 initial configurations are randomly sampled for a hopper (top) and quadruped (bottom). Perturbations from the reference initial configuration include large translations, tilts, and joint angles offsets. For all the samples, and both systems, the policy successfully recovers to the reference gait.}
	\label{monte_carlo}
\end{figure}

\emph{Monte Carlo:}
In order to assess the robustness of CI-MPC, we perform Monte Carlo analysis on two systems: the hopper and planar quadruped. The robots are tasked with tracking a reference gait and we initialize the systems with configurations that are randomly perturbed from the reference trajectory. We use 100 randomly sampled initial conditions for each system; the hopper recovers from significant orientation offsets and the quadruped is robust to large drops (Fig. \ref{monte_carlo}).

\subsection{\diff{Hardware}}
\diff{In this section we present hardware results that demonstrate two capabilities of this work: 1) real-time performance of our algorithm on hardware 2) the robustness of CI-MPC policies to unmodeled disturbances. Three behaviors are executed on a Unitree Go1 quadruped \cite{unitree_go1}: first, a trotting gait that is robust to large external disturbances; second, a non-periodic motion where the system moves towards a wall before balance against it using its front feet; third, the system placing both of its feet onto a step. Videos of the experiments are included with the associated materials.}

\diff{\emph{Point-foot model:}
We utilize a simplified point-foot model of the quadruped that neglects leg dynamics. This model comprises 36 states which include the positions and velocities of the body and each foot. The orientation of the body is represented with Euler angles. The controls are three-dimensional forces applied to each foot, which is modeled as a point mass. Reference motions are generated offline with this model using contact-implicit trajectory optimization \cite{manchester2020variational}.}

\diff{The primary reason for these modeling simplifications is to reduce the online computational requirement when evaluating the dynamics and their derivatives while still being able to reason about new contact sequences online. Importantly, unlike traditional convex quadratic programming policies \cite{bledt2017policy}, which assume a fixed contact sequence for the feet, our model is contact-implicit and enables new foot-step sequences to be generated online.}

\diff{\emph{Experimental setup:} 
A reference trajectory generated offline is tracked online using CI-MPC. A pre-tuned low-level controller generates joint torques that aim to match the forces specified by the policy that should be applied by each leg at the foot.
The CI-MPC policy is written in Julia and is precompiled in order to interface with an existing C++ low-level controller running at $1000$Hz. State feedback to the CI-MPC policy and the low-level controller is provided by a $1000$Hz Kalman filter which utilizes joint encoders, onboard IMU, and external motion-capture tracking to estimate the robot state. The policy, state estimator, and low-level controller run on a computer equipped with an Intel i9-12900KS CPU and 64GB of memory. The joint torque commands are sent to the quadruped via an ethernet connection. Additional information is provided in Table \ref{fig:hardware_timing_table}.}

\diff{\emph{Trotting:}
In this example, the specified behavior is trotting in place (Fig. \ref{quadruped_hardware}). The reference trajectory is 0.8 seconds and is repeated to form a continuous gait. The policy planning horizon is 0.10 seconds and the model uses a time discretization of 0.05 seconds. The policy runs at an average rate 100 Hz.}

\begin{table}[t]
	\centering
    \caption{Comparison between CI-MPC and Pratt state-machine \cite{pratt2000thesis} policies for flamingo system on flat and inclined terrains. We report the number of steps taken by the robot on the flat terrain and compare the maximum incline traversed by our policy in simulation with reported results$^{\dagger}$ \cite{pratt1998intuitive}.}
    
	\begin{tabular}{c c c}
		\toprule
	    \textbf{Policy} &
		\textbf{Flat} &
		\textbf{Incline}\\
		\toprule
		CI-MPC & $+100\phantom{^{\dagger}}$ & \boldmath $10$ \unboldmath \textbf{deg}. \\ 
		Pratt & $+100^{\dagger}$ &  $ 5^{\dagger}$  deg. \\ 
		\toprule
	\end{tabular}
    \label{fig:biped_results}
    \vspace{-5mm}
\end{table}

\diff{\emph{Wall stand:}
In this example, the specified behavior is transitioning from four feet on the ground to standing against the wall with two feet (Fig. \ref{quadruped_hardware}). The reference trajectory is 19.75 seconds. The policy planning horizon is 0.15 seconds and the model uses a time discretization of 0.05 seconds. The policy runs at an average rate of 100 Hz.}

\diff{\emph{Step:}
In this example, the specified behavior has the quadruped place its right foot onto a step, followed by its left foot (Fig. \ref{quadruped_hardware}). The entire reference trajectory is 7.0. The policy planning horizon is 0.1 seconds and the model uses a time discretization of 0.05 seconds. The policy runs at an average rate of 100 Hz.}

\section{Conclusion} \label{conclusion}

\diff{CI-MPC is capable of generating dynamic behaviors by robustly tracking reference trajectories through contact despite disturbances, model mismatch, and uncertain environments. In this section we conclude with a discussion of limitations and directions for future research.}

\diff{\subsection{Limitations}
We highlight important limitations including: approximations, contact model, and reliability, that should be considered before deploying CI-MPC policies.}

\diff{\emph{Approximations:} 
For the online trajectory optimization problem, strategically approximated contact dynamics are utilized for planning.} This enables expensive gradient computations and partial matrix factorizations to be performed in an offline stage, in order to substantially reduce online computation. \diff{In the examples, we find that short planning horizons, typically between $0.1$ and $0.25$ seconds, are sufficient. Despite the approximations introduced by these simplifications, in practice, the controls optimized for the fast contact dynamics work well in simulation and on hardware. However, it remains to be seen how well these approximations work for control of highly dynamic behaviours or scenarios that require longer planning horizons, particularly when deployed on hardware.}

\diff{\emph{Contact model:} 
The physics of hard contact produces non-smooth and discontinuous gradients. With our custom interior-point method for the contact dynamics solver, we can efficiently compute smooth gradients in a principled way by exploiting intermediate results, parameterized by the central-path parameter. Hard contact is simulated by returning results from the contact dynamics solver with a central-path value $\kappa_{\mbox{sim}} = 1\mbox{e-}6$, whereas gradients are computed using intermediate results from the solve parameterized by $\kappa_{\mbox{grad}} \approx 1\mbox{e-}4$.}

During online optimization, we prioritize fast updates by solving the trajectory-tracking problem to coarse tolerances. In this context, imposing highly accurate contact physics would be wasteful in terms of computational resources. \diff{As a result, the central-path value for the planning dynamics is fixed to the gradient central-path value in order to reduce online computation. This selection was made to balance capturing accurate physics with producing usefully smooth gradients. Empirically, we observe that using these dynamics with slightly soften contact dynamics enhanced the convergence of the trajectory-tracking solver and likely enables the policy to more easily discover new contact sequences. Importantly, the allowed interpenetration with these tolerances is sub-millimeter---much less than allowed by MuJoCo's default settings---but, this raises the question, is simulation of perfectly hard contact actually necessary, or useful, for reliable planning and control of non-smooth systems?}

\begin{table}[t]
	\centering
    \caption{\diff{The CI-MPC policy runs at real-time rates meaning that the time required to compute the control is always smaller than the reference time step and the policy is able to successfully track the specified trajectory. Experiments are run on a computer equipped with an Intel Core i9-9900 CPU and 32GB of memory.}}
	\begin{tabular}{c c c c}
		\toprule
	    \textbf{System} &
		\textbf{Planning Horizon} &
		\textbf{Time Step} & 
        \textbf{Real-Time} \\
		\toprule
		pushbot          & $1.60$ s & $0.04$ s & \checkmark \\ 
		hopper (2D)      & $0.10$ s & $0.01$ s & \checkmark \\ 
		hopper (3D)      & $0.20$ s & $0.01$ s & \checkmark \\ 
		quadruped        & $0.16$ s & $0.016$ s & \checkmark \\
		biped            & $0.23$ s & $0.016$ s & \checkmark \\
		\toprule
	\end{tabular}
    \label{fig:timing_table}
\end{table}

\diff{\emph{Reliability:}
Generating high-quality reference trajectories is crucial for CI-MPC. Contact-implicit trajectory optimization \cite{posa2014direct, manchester2020variational} is a powerful tool for generating these trajectories. However, it is notorious for poor convergence properties, despite relying on robust large-scale constrained solvers for non-convex problems and even good warm starting. This unreliability makes online optimization generally impractical---motivating this work. Ultimately, for CI-MPC to be of practical value, generation of references trajectories, even in the offline setting, must be improved. This may be possible with specialized solvers for contact-implicit trajectory optimization \cite{howell2022calipso}, alternative rollout-based methods that leverage the reliability of one-step contact dynamics \cite{howell2022trajectory}, or learning-based approaches \cite{heess2017emergence}.}

\diff{Additionally, our hardware experiments demonstrate the real-time capabilities of CI-MPC and its ability to be robust in many scenarios. However, in practice, we find that the classic convex MPC policy \cite{di2018dynamic} is significantly more robust for general locomotion because that policy has an additional online foothold selection strategy (the Raibert strategy). Adding this strategy to CI-MPC will likely increase its speed and reliability to match the performance of the baseline convex MPC policy. Performance will likely be improved further with a more efficient C/C++ implementation that utilizes multi-threading for parallel evaluations of the fast contact dynamics.}

\begin{table}[t]
	\centering
    \caption{\diff{Hardware experiments with CI-MPC tracking different reference trajectories on a Unitree Go1 quadruped.}}
    \diff{
	\begin{tabular}{c c c c}
        \toprule
        \textbf{Settings}  & \textbf{Trotting} & \textbf{Wall} & \textbf{Step}\\
        \toprule
	    reference trajectory length & $0.8$ s & $19.75$ s & $7.0$ s \\
        reference time step & $0.05$ s & $0.05$ s & $0.05$ s\\
		policy planning horizon & $0.1$ s & $0.15$ s  & $0.1$ s\\
		policy rate & $100$ Hz & $100$ Hz & $100$ Hz\\
		\toprule
	\end{tabular}
    }
    \label{fig:hardware_timing_table}
\end{table}

\diff{\subsection{Future Work}
In summary, we have presented fast differentiable contact dynamics that can be utilized in an MPC framework that performs robust tracking for robotic systems that make and break contact with their environments. There remain many exciting avenues to explore in future work. First, it should be possible to perform higher-fidelity convex approximations of the contact dynamics that utilize second-order friction cones instead of its linearized approximation. This could enable tracking of highly dynamic behaviors that leverage accurate sliding contacts. Second, a natural extension of this work, which was focused on locomotion, is to the manipulation domain, potentially with quasi-static models, where control through contact is similarly an open problem, but where dynamics are slower and more amenable to online optimization. Third, in our hardware experiments, we utilized a simplified planning model for the quadruped. Similar point-contact models should extend to bipeds and humanoid systems, potentially even dexterous hands. Lastly, a library of template behaviors, comprising CI-MPC policies, could be composed to enable more diverse behavior online with a high-level agent composing templates in a task-and-motion-planning framework in order to generate complex long-horizon plans.}

\begin{figure*}[t]
	\centering
    \includegraphics[width=1.00\textwidth, height=3.00cm]{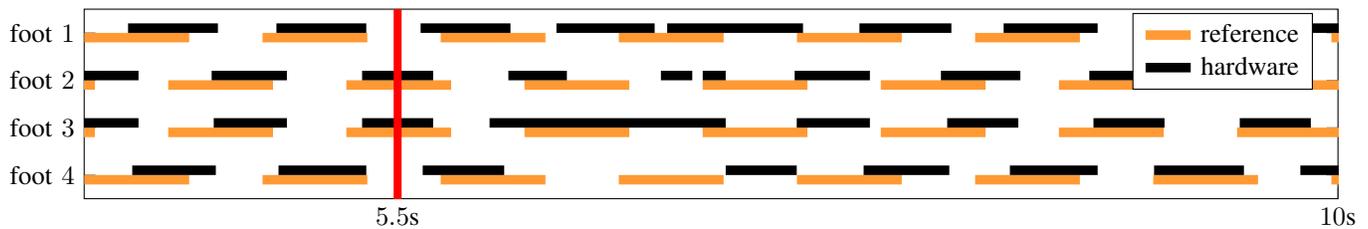}		\caption{diff{Contact sequence for quadruped trotting. An external disturbance is applied to the system (red) and the policy is able to generate a new contact sequence online (black) that differs significantly from the reference plan (orange).}}
	\label{contact_sequence}
\end{figure*}

\section*{Acknowledgments}

This work was supported in part by Frontier Robotics, Innovative Research Excellence, Honda R\&D Co., Ltd, ONR award N00014-18-1-2830, NSF NRI award 1830402, and DARPA YFA award D18AP00064. Toyota Research Institute (``TRI") provided funds to assist the authors with their research but this article solely reflects the opinions and conclusions of its authors and not TRI or any other Toyota entity. 

\bibliographystyle{IEEEtran}
\bibliography{main}

\clearpage 

\begin{IEEEbiography}
    [{\includegraphics[width=1in,height=1.25in,clip,keepaspectratio]{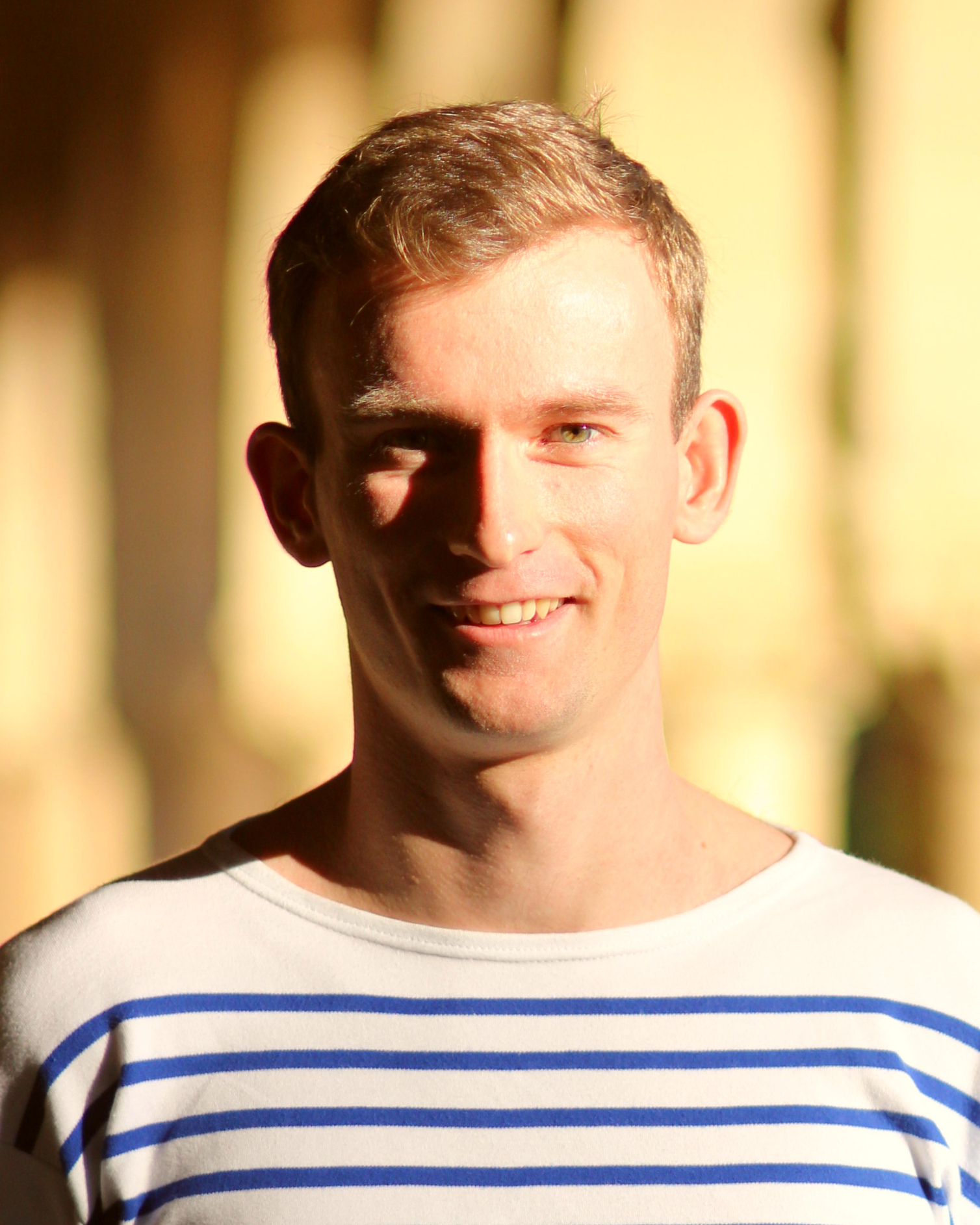}}]{Simon Le Cleac'h} is a graduate student with the Robotic Exploration Lab at Carnegie Mellon University and with the Multi-Robot Systems Lab at Stanford University. He received his BS in Engineering from Ecole Centrale Paris in 2016 and his MS in Mechanical Engineering from Stanford University in 2019. His research interests include differentiable optimization for control and simulation of robotics systems.
\end{IEEEbiography}

\begin{IEEEbiography}
    [{\includegraphics[width=1in,height=1.25in,clip,keepaspectratio]{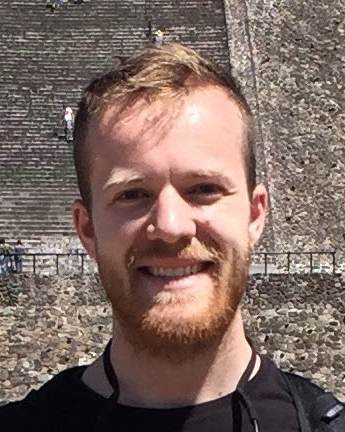}}]{Taylor Howell} is a graduate student at Stanford University and with the Robotic Exploration Lab. He received his BS in mechanical engineering in 2016 from the University of Utah and his MS in mechanical engineering from Stanford University in 2019. His research interests include numerical optimization and control for robotic systems.
\end{IEEEbiography}

\begin{IEEEbiography}
    [{\includegraphics[width=1in,height=1.25in,clip,keepaspectratio]{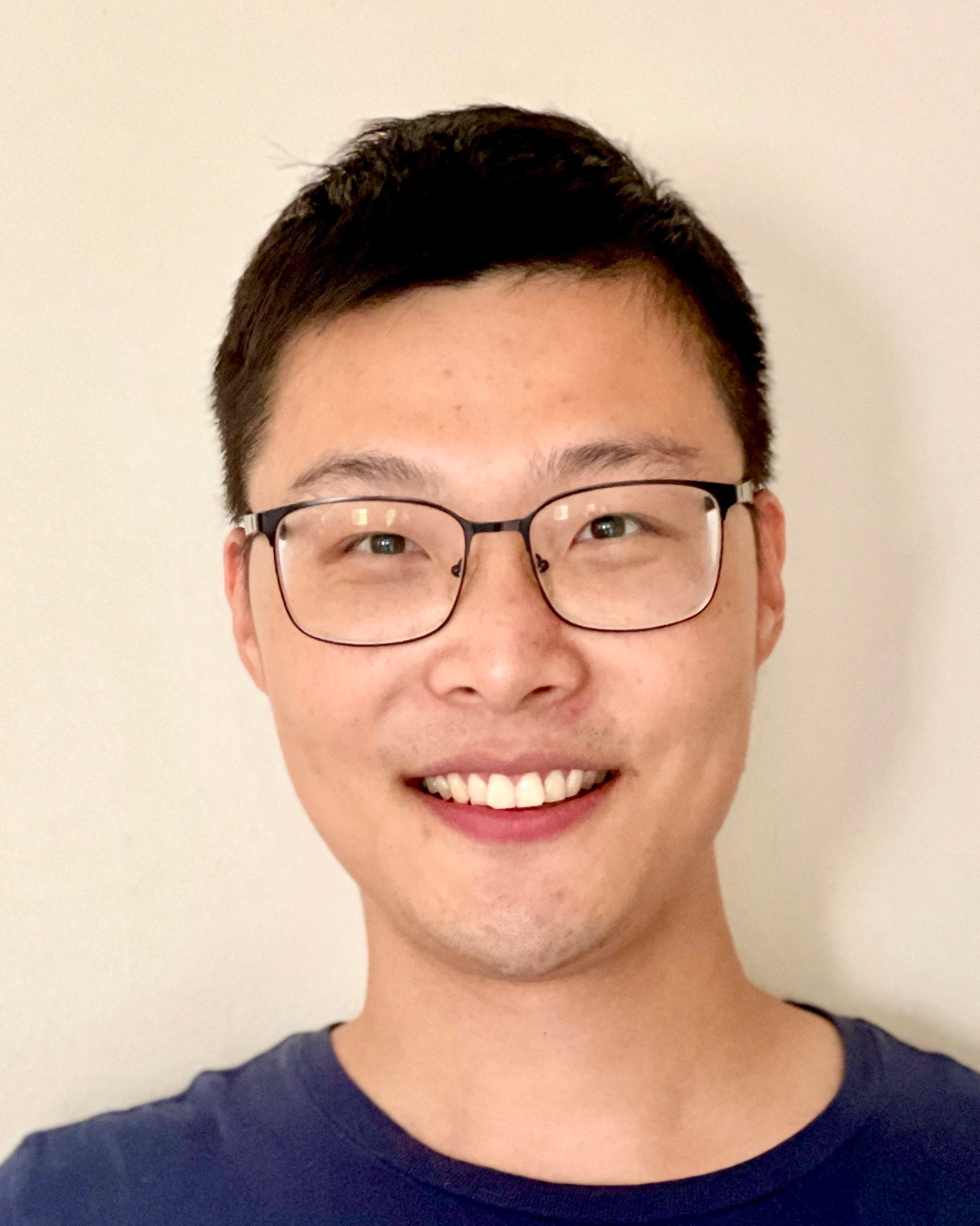}}]{Shuo Yang} is a graduate student at the Robotics Exploration Lab at Carnegie Mellon University. He received his BEng degree in Computer Engineering in 2012 and MPhil degree in Electrical and Computer Engineering in 2015, both from the Hong Kong University of Science and Technology. His research interests include trajectory optimization and state estimation for robotic systems.
\end{IEEEbiography}

\begin{IEEEbiography}[{\includegraphics[width=1in,height=1.25in,clip,keepaspectratio]{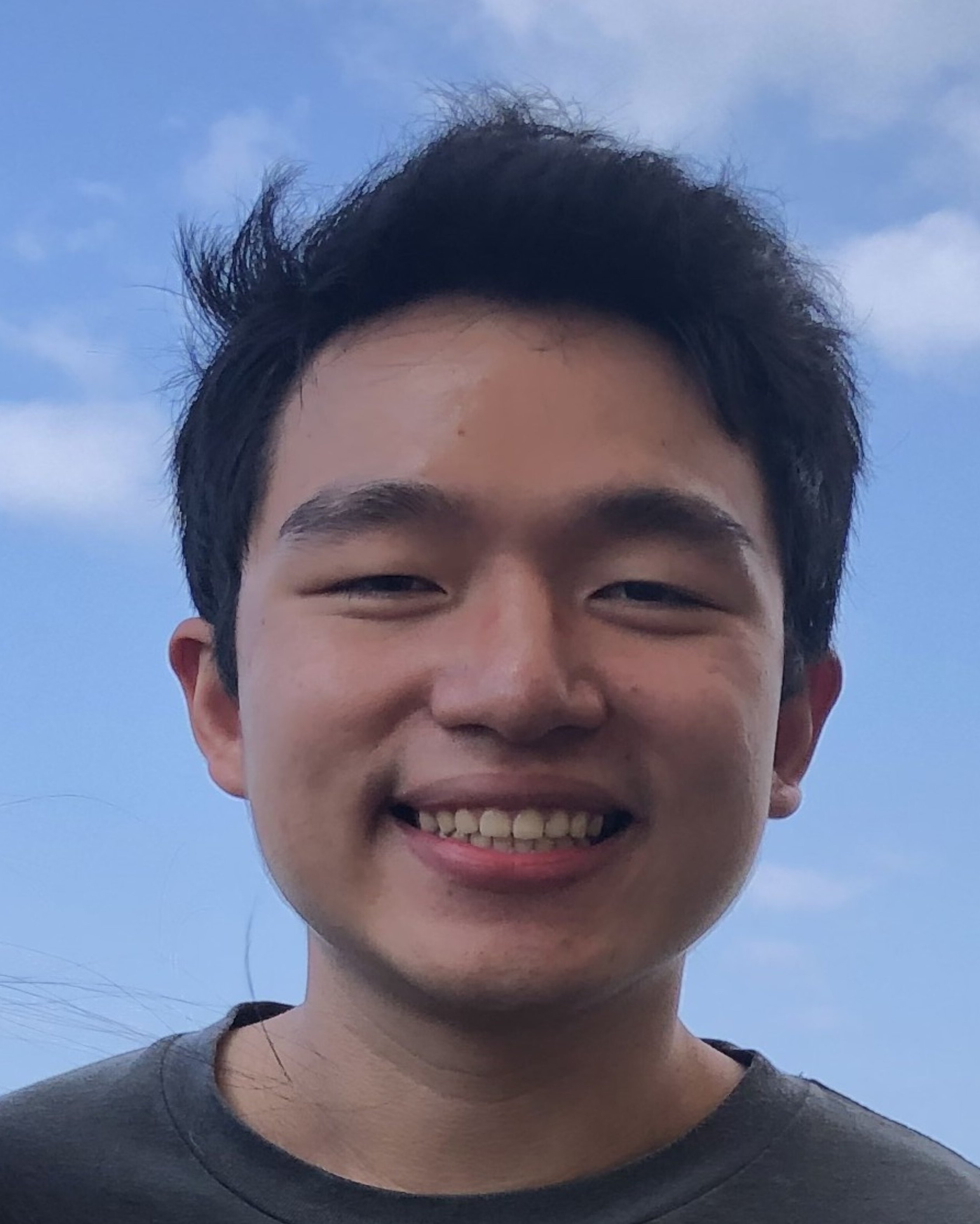}}]{Chi-Yen Lee} is a recent graduate from the Robotics Exploration Lab at Carnegie Mellon University. He received his BS in Engineering from Harvey Mudd College in 2019 and his MS in Robotics from Carnegie Mellon University in 2022. His research interests include optimal control and state estimation for robotics system.
\end{IEEEbiography}

\begin{IEEEbiography}[{\includegraphics[width=1in,height=1.25in,clip,keepaspectratio]{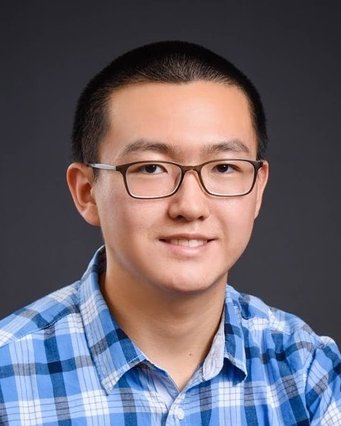}}]{John Zhang} is a graduate student at the Robotic Exploration Lab at Carnegie Mellon University. He received his B.S. in Mechanical Engineering and a minor in Computer Science from the Georgia Institute of Technology. His research interests include optimization and control for robotic systems.
\end{IEEEbiography}

\begin{IEEEbiography}[{\includegraphics[width=1in,height=1.25in,clip,keepaspectratio]{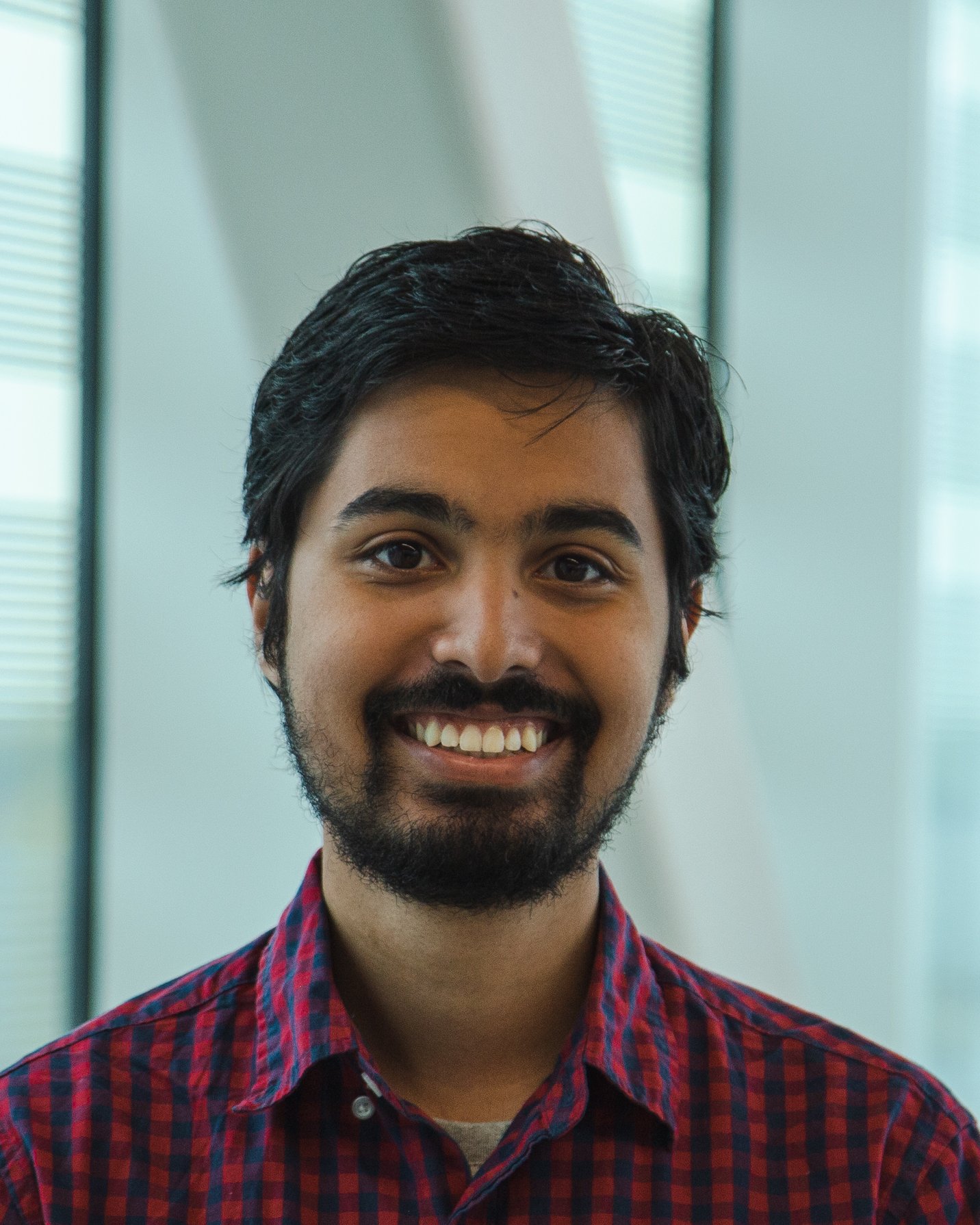}}]{Arun Bishop} is a Phd Student in the Robotics Institute at Carnegie Mellon University.
He received his B.S.E. in Mechanical Engineering and his M.S. in Robotics from the University of Michigan working on
robust bipedal walking. His research interests include optimal control and design of legged robotic systems for applications
in exploration and search and rescue.
\end{IEEEbiography}

\begin{IEEEbiography}   
    [{\includegraphics[width=1in,height=1.25in,clip,keepaspectratio]{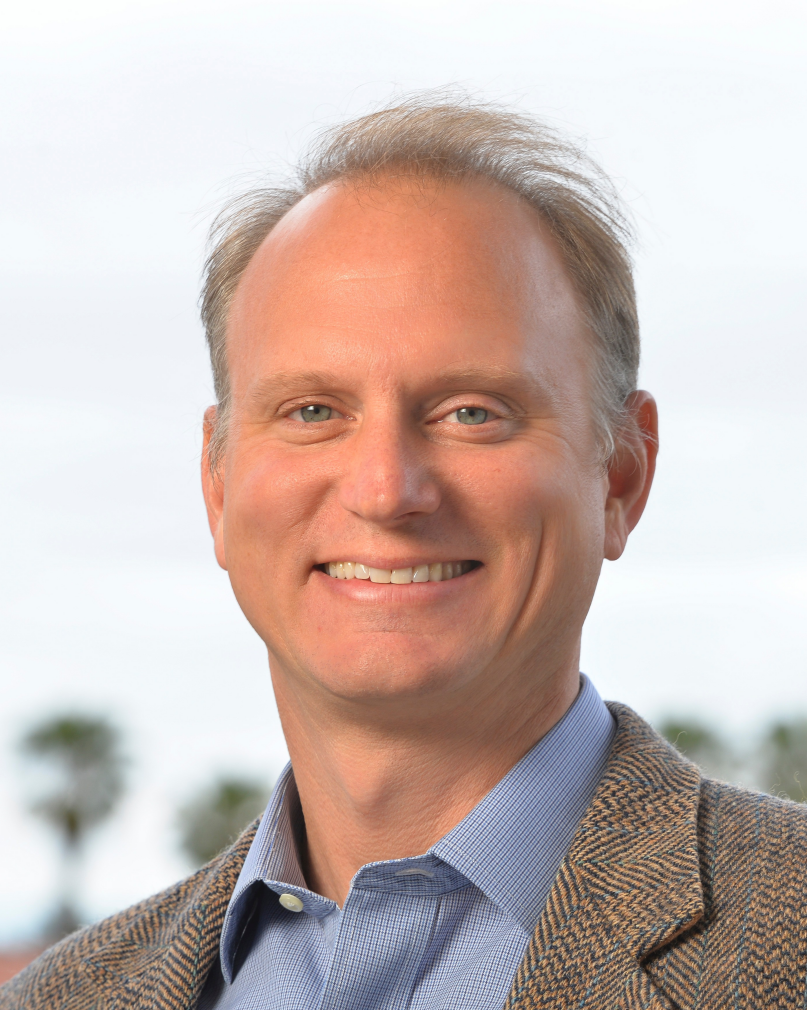}}]{Mac Schwager} is an assistant professor with the Aeronautics and Astronautics Department at Stanford University.  He obtained his BS degree in 2000 from Stanford University, his MS degree from MIT in 2005, and his PhD degree from MIT in 2009.  He was a postdoctoral researcher working jointly in the GRASP lab at the University of Pennsylvania and CSAIL at MIT from 2010 to 2012, and was an assistant professor at Boston University from 2012 to 2015.  He received the NSF CAREER award in 2014, the DARPA YFA in 2018, and a Google faculty research award in 2018, and the IROS Toshio Fukuda Young Professional Award in 2019.  His research interests are in distributed algorithms for control, perception, and learning in groups of robots, and models of cooperation and competition in groups of engineered and natural agents.
\end{IEEEbiography}

\begin{IEEEbiography}
    [{\includegraphics[width=1in,height=1.25in,clip,keepaspectratio]{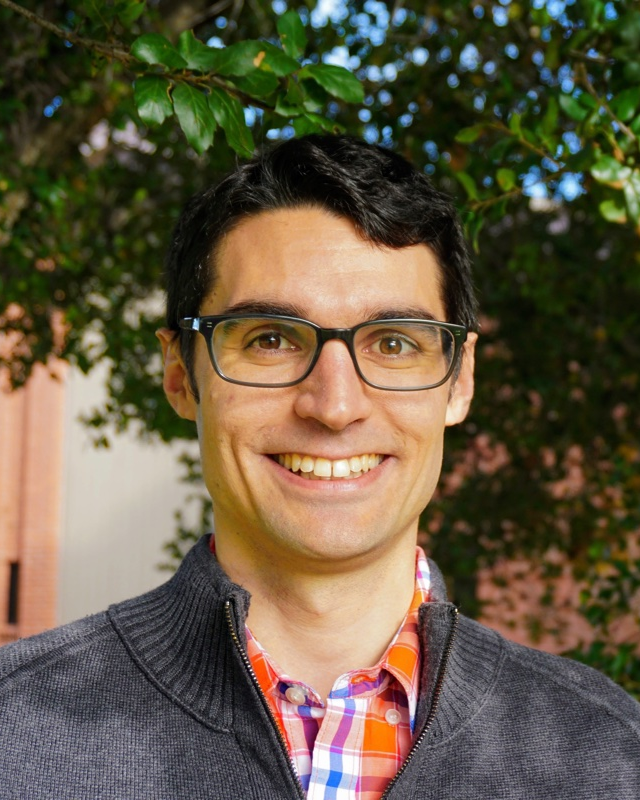}}]{Zachary Manchester} is an assistant professor in the Robotics Institute at Carnegie Mellon University and founder of the Robotic Exploration Lab. He received a PhD in aerospace engineering in 2015 and a BS in applied physics in 2009, both from Cornell University. He was a postdoctoral fellow in the Agile Robotics Lab at Harvard from 2015 to 2017 and an assistant professor at Stanford from 2018 to 2020. He received the NASA Early Career Faculty Award in 2018 and a Google Faculty Research Award in 2020. His research interest include numerical optimization, control and estimation with applications to aerospace and robotic systems with challenging nonlinear dynamics.
\end{IEEEbiography}

\end{document}